%% file: main.tex
\documentclass[twoside]{article}

\input{preamble}
\usepackage[accepted]{icml2024}

\begin{document}

\twocolumn[
\icmltitle{
PASOA- PArticle baSed {Bayesian} Optimal Adaptive design
}
\icmlsetsymbol{equal}{*}

\begin{icmlauthorlist}
\icmlauthor{Jacopo Iollo}{InriaG,InriaS,Cerema}
\icmlauthor{Christophe Heinkelé}{Cerema}
\icmlauthor{Pierre Alliez}{InriaS}
\icmlauthor{Florence Forbes}{InriaG}

\end{icmlauthorlist}

\icmlaffiliation{InriaG}{Université Grenoble Alpes, Inria, CNRS, G-INP, France}
\icmlaffiliation{Cerema}{Cerema, Endsum-Strasbourg, France}
\icmlaffiliation{InriaS}{Université C\^ote d'Azur, Inria, France}

\icmlcorrespondingauthor{Jacopo Iollo}{jacopo.iollo@inria.fr}
\icmlcorrespondingauthor{Florence Forbes}{florence.forbes@inria.fr}

\icmlkeywords{Bayesian Experimental Design, Sequential Monte Carlo, Tempering, Stochastic Optimization}
\vskip 0.3in
]

\printAffiliationsAndNotice{}

\begin{abstract}
We propose a new procedure {named PASOA}, for Bayesian experimental design, that performs sequential design optimization by simultaneously providing accurate estimates of successive posterior distributions for parameter inference. The sequential design process is carried out via a contrastive estimation principle, using stochastic optimization and Sequential Monte Carlo (SMC) samplers to maximise the  Expected Information Gain (EIG).  As larger information gains are obtained for larger distances between successive posterior distributions, this EIG objective may worsen classical SMC performance.
To handle this issue, tempering is proposed to have both a large information gain and an accurate SMC sampling{, that we show is crucial for performance}.
This novel combination of stochastic optimization and  tempered SMC  allows to jointly handle design optimization and  parameter inference. We provide a proof that the obtained optimal design estimators benefit from some consistency property. Numerical experiments confirm the potential of the approach, which outperforms other recent existing procedures.
\end{abstract}

\section{Introduction}
\begin{figure*}[hpt!]
  \centering
    \includegraphics[trim={.3cm 0.7cm 0.2cm 0},clip, height=2.06cm, width=.17\linewidth]{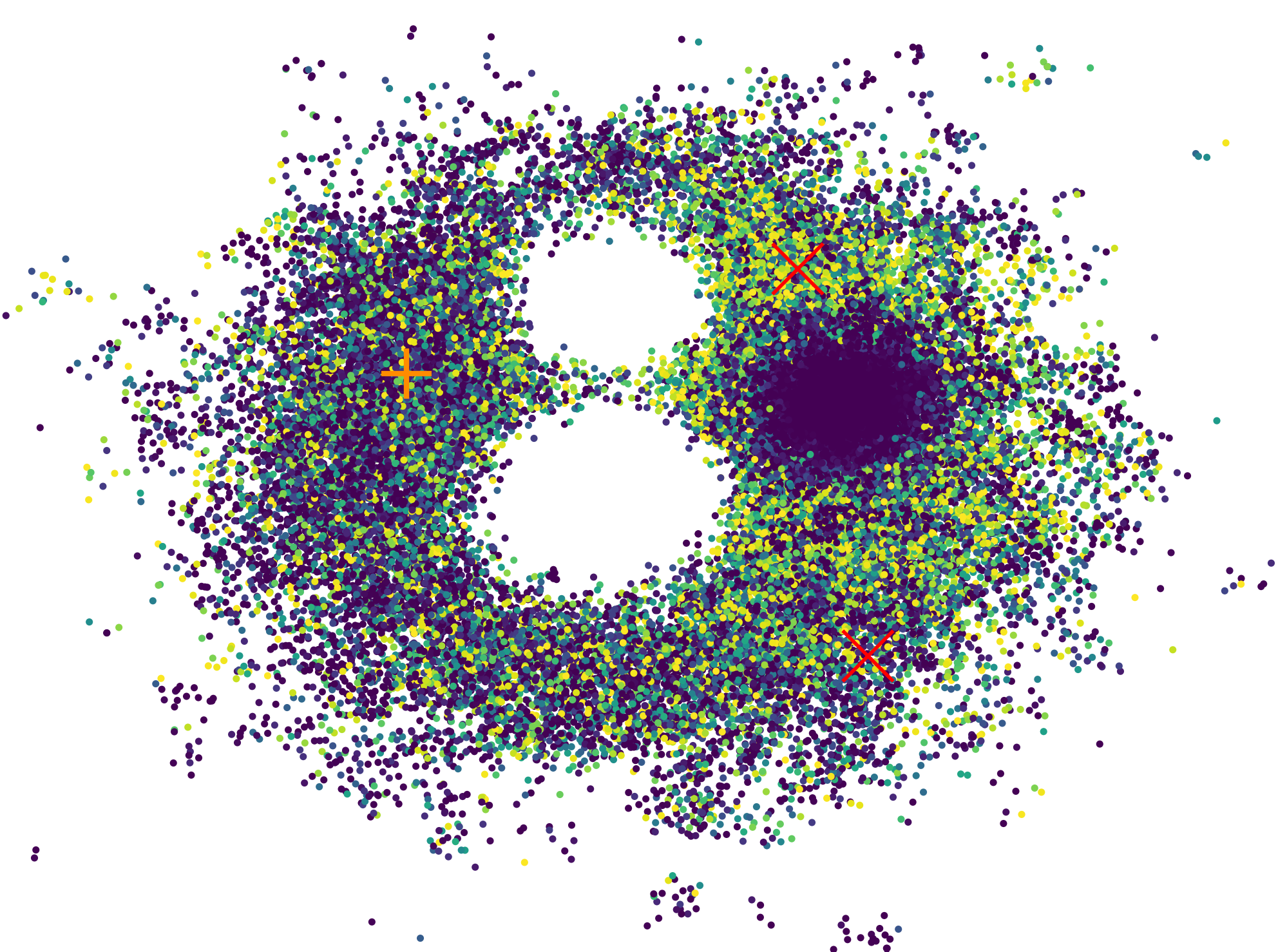} \hspace{0.35cm}
    \includegraphics[trim={.3cm 0.7cm 0.2cm 0},clip, height=2.06cm, width=.17\linewidth]{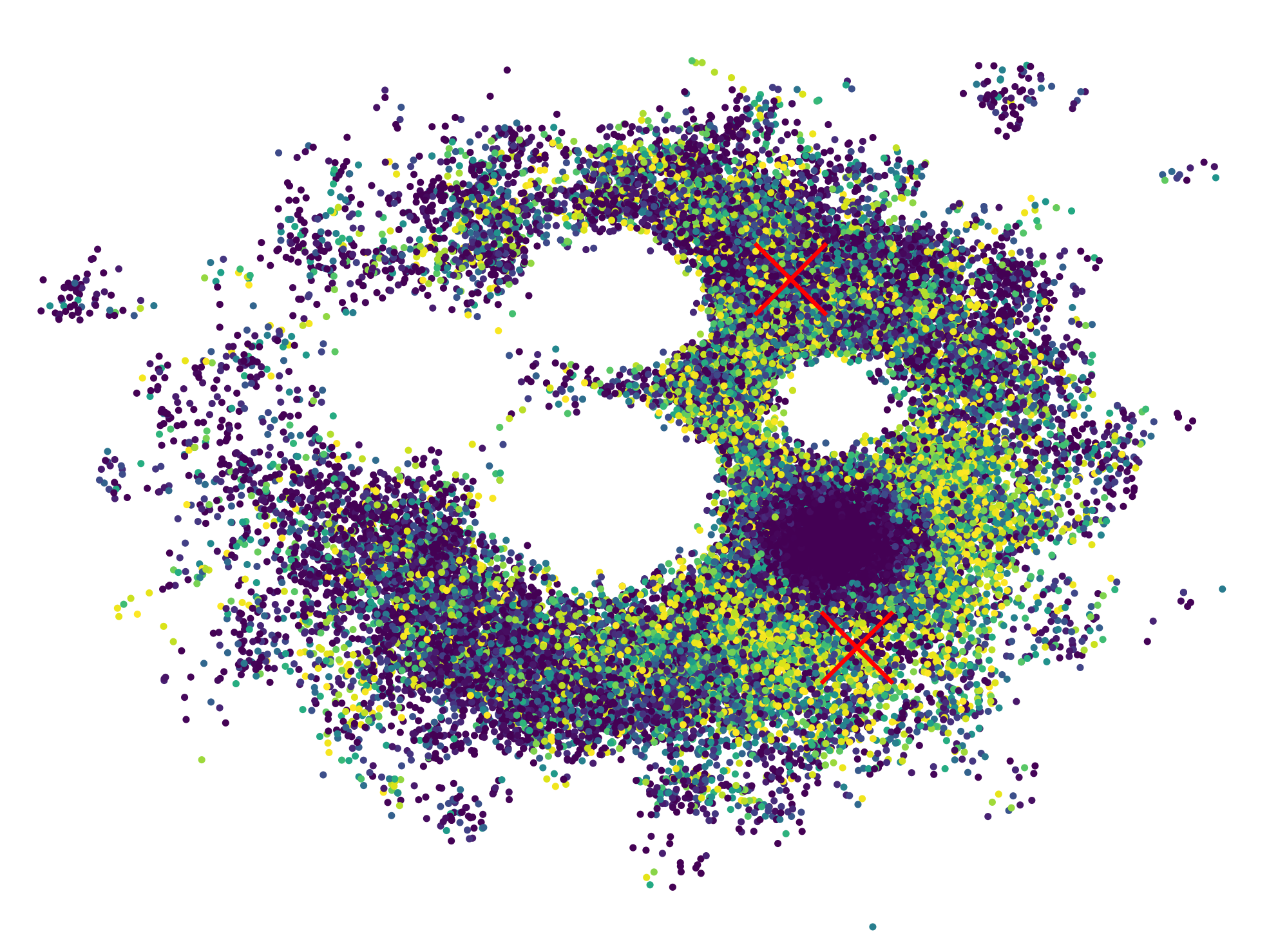} \hspace{0.35cm}
    \includegraphics[trim={.3cm 0.7cm 0.2cm 0},clip, height=2.06cm, width=.17\linewidth]{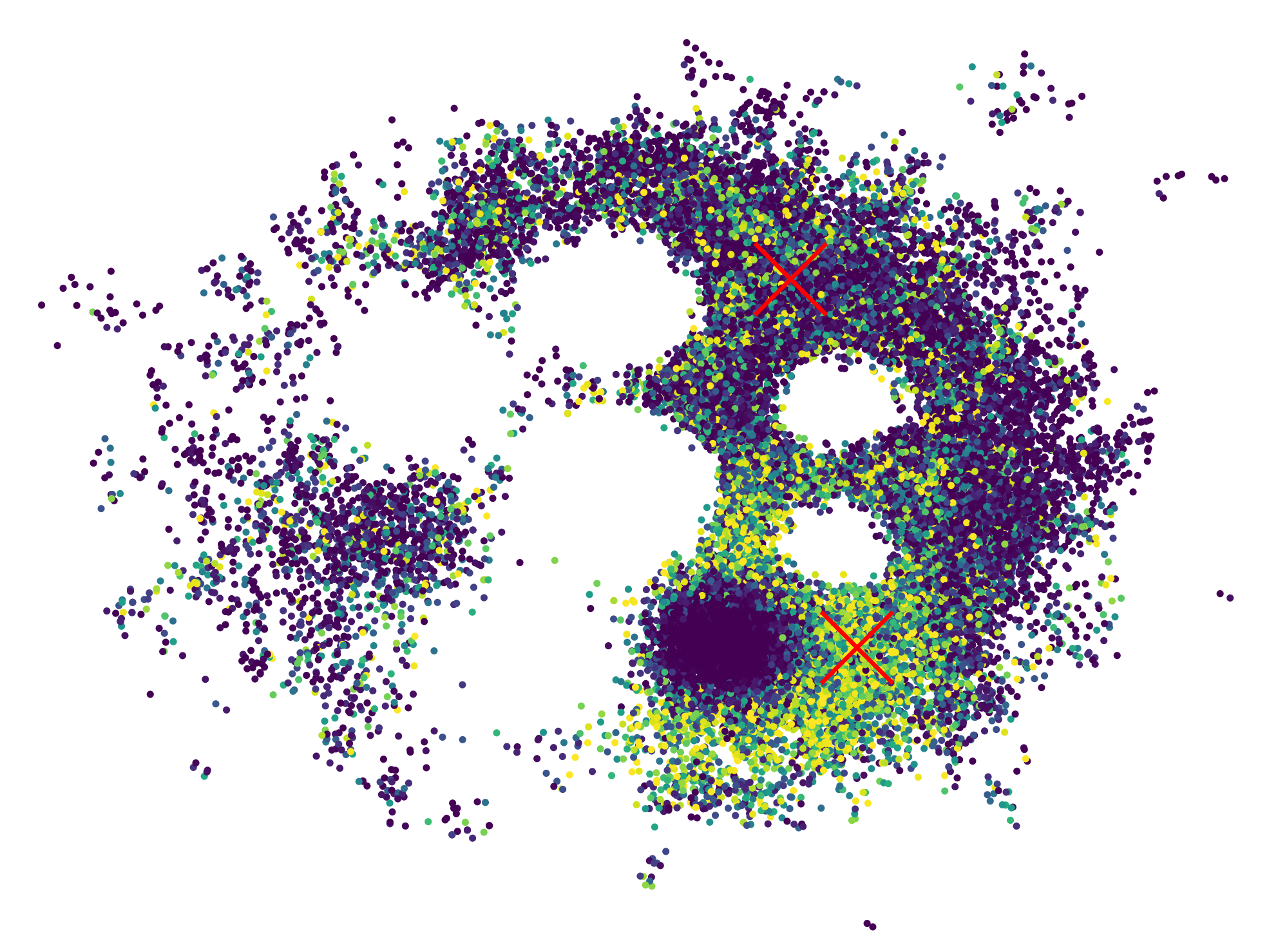} \hspace{0.35cm}
    \includegraphics[trim={.3cm 0.7cm 0.2cm 0},clip, height=2.06cm, width=.17\linewidth]{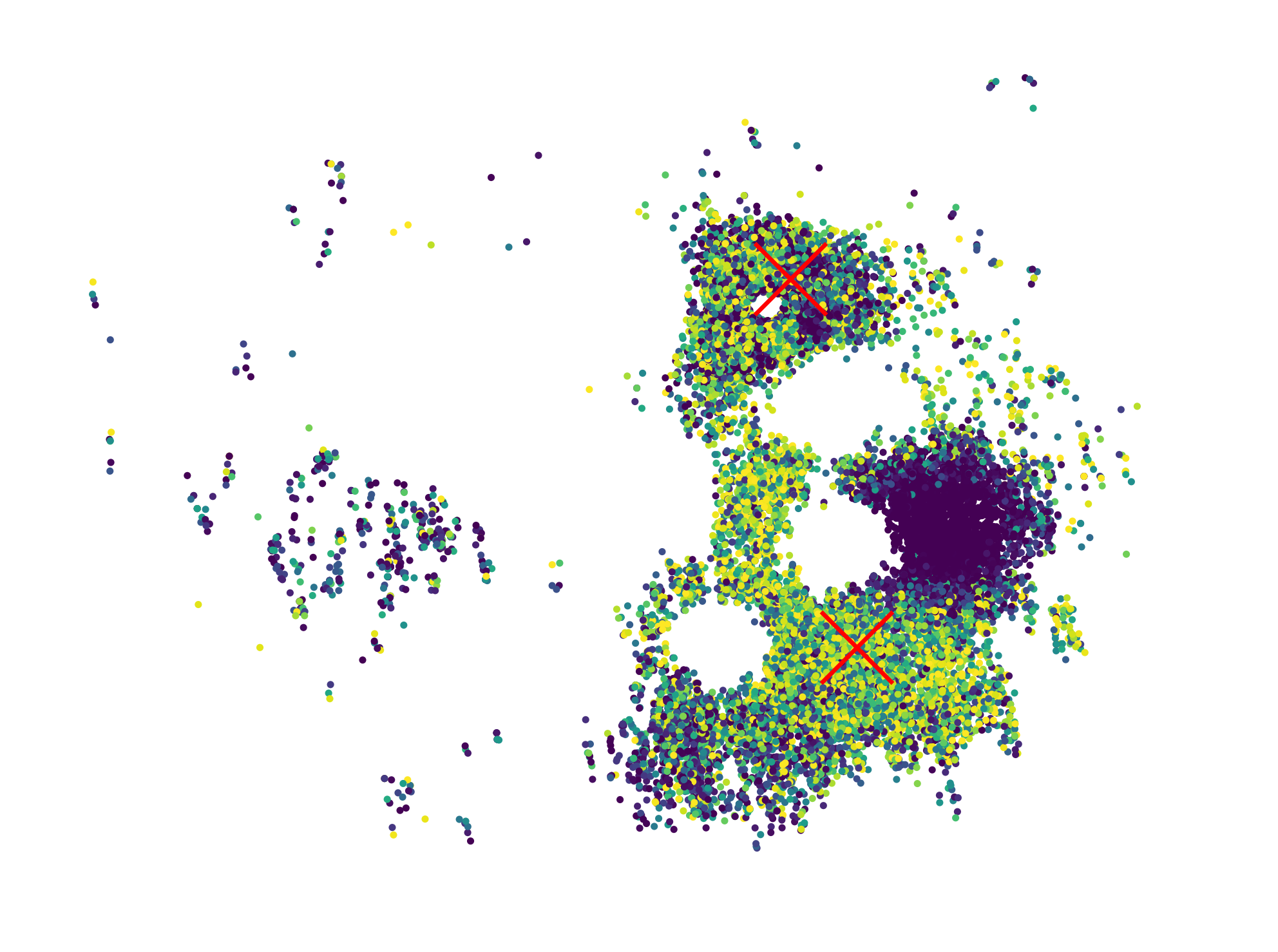}\\
    \includegraphics[trim={.3cm 0.7cm 0.2cm 0},clip, height=2.06cm, width=.17\linewidth]{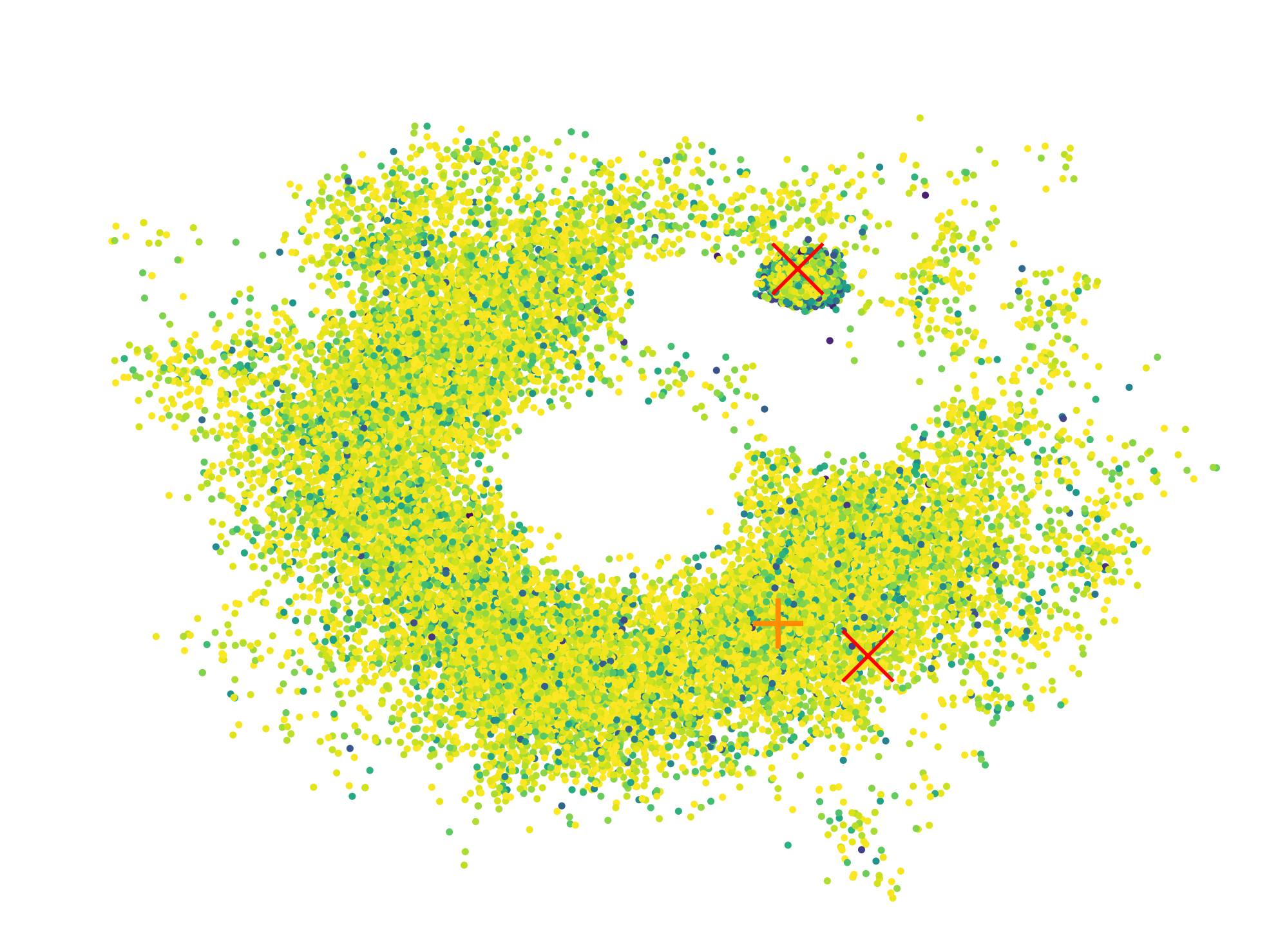}\hspace{0.35cm}
    \includegraphics[trim={.3cm 0.7cm 0.2cm 0},clip, height=2.06cm, width=.17\linewidth]{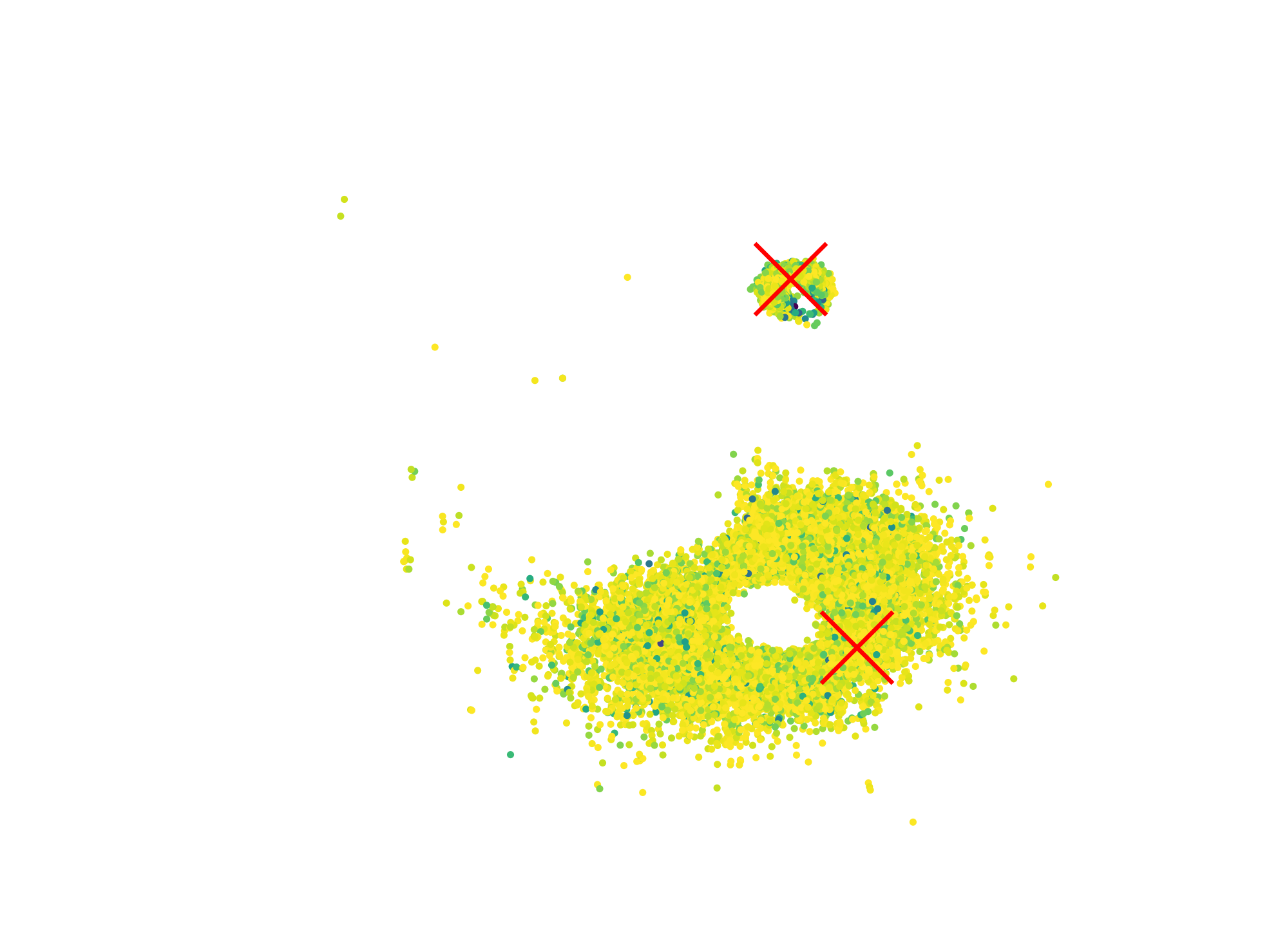}\hspace{0.35cm}
    \includegraphics[trim={.3cm 0.7cm 0.2cm 0},clip, height=2.06cm, width=.17\linewidth]{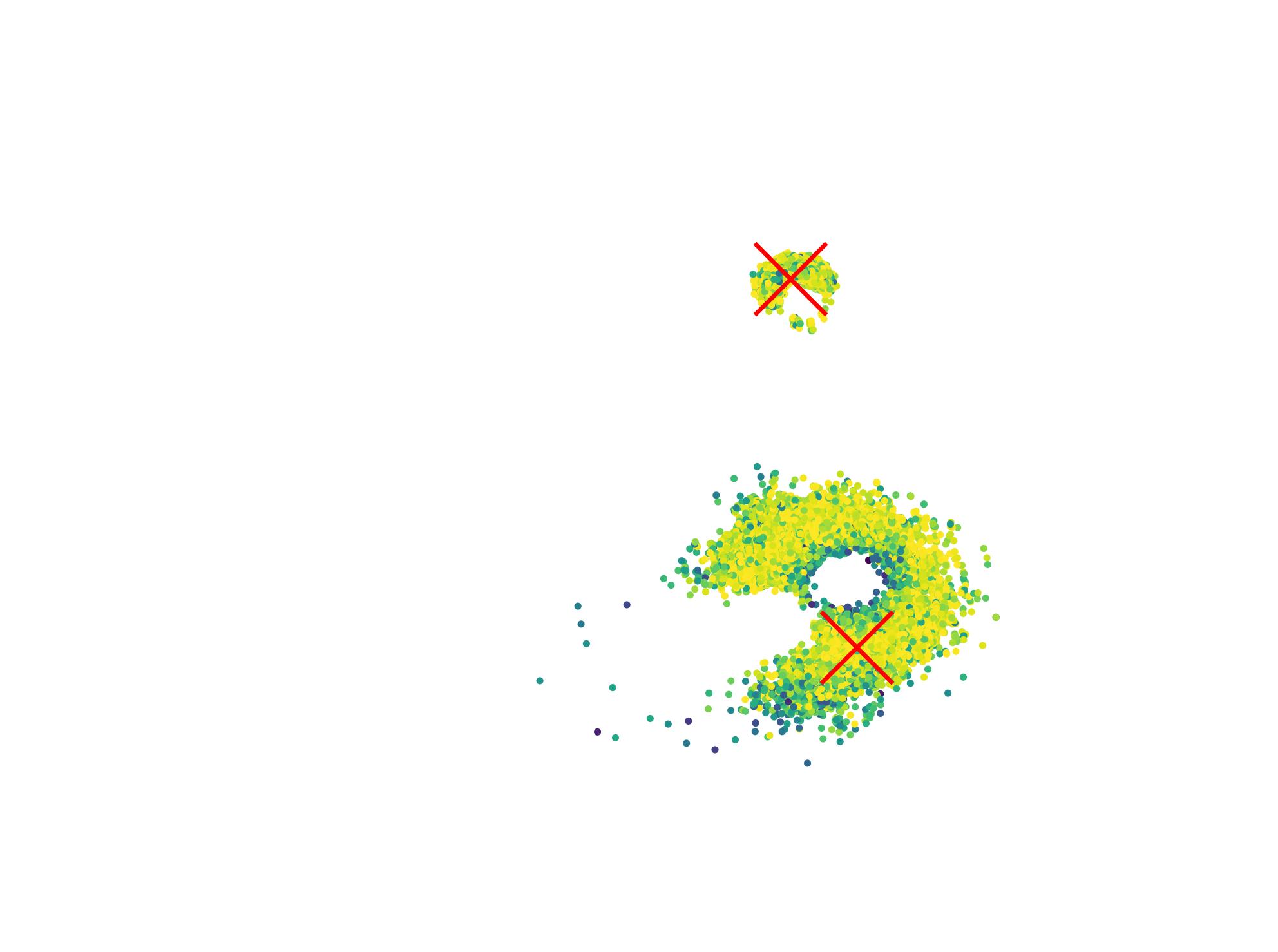}\hspace{0.35cm}
    \includegraphics[trim={.3cm 0.7cm 0.2cm 0},clip, height=2.06cm, width=.17\linewidth]{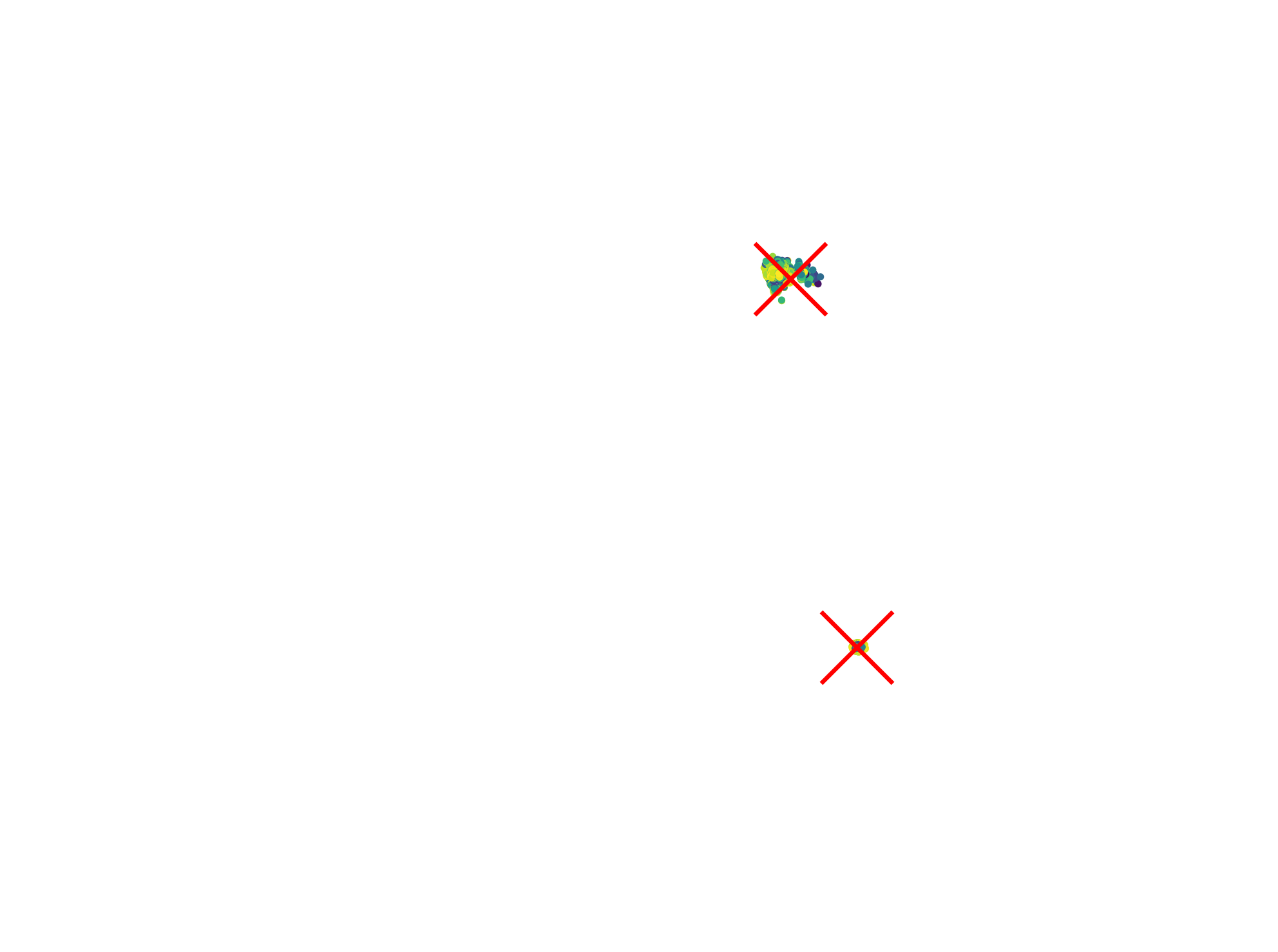}
    \caption{\footnotesize Source location example at steps 5, 7, 9, 11. Over design steps, particles concentrate  faster to true sources (red crosses) with PASOA (2nd line) than with SMC (1st line). Lower particle weights in blue, higher in yellow.
    }
    \label{fig:left4}
\end{figure*}

A design refers to some experimental conditions required to perform an experiment and get observations from the phenomenon under study.
Assuming that such a design is characterized by some parameters denoted by $\xib$, simple examples of $\xib$ are coordinates in a 2D space  or a frequency at which we wish to measure a quantity of interest.
The overall goal of experimental design can be summarized as the acquisition of good quality data, which is of increasing importance in numerous scientific and industrial applications, see {\it e.g.} \cite{Ryan2016review,Rainforth2023}. Quality can be defined from numerous view points, allocating resources for information gathering, improving precision and/or prediction or reducing experimental costs.
In this work, we assume  that the desired designs are continuous parameters $\xib~\in~{\cal E}$ that maximize information on some parameters of interest $\thetab \in \Thetab \subset \Rset^m$. In Bayesian Optimal Experimental Design (BOED), a Bayesian model is specified over
$\thetab$ \cite{Chaloner1995,Sebastiani2000,Amzal2006,Ryan2016review,Rainforth2023}. The criterion defining the optimality of a design may then depend on the application \cite{Ryan2016review,Kleinegesse2021} but the most used of them is the expected information gain (EIG). It targets designs that most reduce the entropy of the posterior distribution over $\thetab$, or equivalently that maximize the Kullback-Leibler divergence between the posterior and the prior, and this
in expectation over all possible experimental outcomes, see eq.~(\ref{def:I},\ref{def:EIGpost}). The Bayesian formulation is particularly convenient in sequential contexts, where we wish to  exploit the results of previous  experiments to guide the design of future ones.
Unfortunately, computing and
optimizing the EIG sequentially requires  sampling  from an evolving sequence of intractable posterior distributions.
We propose  to use  a sequential Monte-Carlo (SMC) approach for an efficient  sequential sampling \cite{del2006sequential,Doucet2018,Naesseth2019,chopin2020introduction,dai2022invitation} and
in the spirit of some original and recent approaches \cite{Huan2014,Kleinegesse2021,Foster2019,Foster2020,Foster2021,Blau2022}, we  adopt a stochastic optimization approach to optimize the EIG. The originality of our approach is to use  SMC  in conjunction with stochastic optimization to efficiently sample the relevant quantities and estimate the required noisy gradients.
However, naive SMC is potentially problematic. In sequential EIG-based optimization, larger gains are obtained for larger  distances between successive posterior distributions, which is a particularly bad scenario for SMC performance.
A solution is to consider
 {\it distribution tempering or annealing}
 %\cite{neal2001,chopin2020introduction,Naesseth2019,Syed2019},
 %% see also biblio notes p.33 in Chopin's book
 \cite{Neal2001,del2006sequential},
 which creates a path of intermediate closer distributions for a more accurate SMC sampling (Figures \ref{fig:left4} and \ref{fig:tempering}).
We name our approach PASOA, for Particle baSed Bayesian Optimal Adaptive design, where particles refer to samples featuring an evolving Bayesian posterior information that adaptively guides the successive design estimation.
To our knowledge, such a combination, of stochastic optimization and SMC or tempered SMC, has not been proposed before for BOED.
It allows to benefit from both techniques' advantages.
Stochastic optimization provides a more scalable way to handle sequential design optimization, while the principled SMC framework allows to efficiently perform an accurate parameter inference by reusing samples from past experiments. This high quality inference is exploited in turn via stochastic optimization to yield informative  design estimators coming with good theoretical features.
We show that  these estimators are consistent and converge in probability to the optimal designs.
On the numerical side, we verify their practical performance on benchmark  design problems.  We observe  significantly higher information gains than previous methods, including sophisticated and reinforcement learning approaches \cite{Blau2022,Foster2020}, with these gains leading, in turn, to improved posterior inference and parameter estimation.
{We thus provide the first theoretical result of this kind in the BOED literature, while also highlighting tempering  as a mean to  greatly enhance the performance of SMC in BOED. }

\section{Related work}

Simulations required for EIG optimization can be obtained  via Markov Chain Monte Carlo (MCMC) algorithms, {see {\it e.g.} \citet{Kleinegesse2020ICML} for a static design.} In a sequential design context,  the necessity to do so at each step is very simulation intensive \cite{Amzal2006,Kleinegesse2020}. {Typically, \citet{Kleinegesse2020} extend \citet{Kleinegesse2020ICML} to a sequential setting but shows up to only $K=4$ or 5 experiments.} As an alternative, SMC samplers have been used in previous work, {\it e.g.} \citet{Drovandi2014,Drovandi2013}  use SMC but only for finite-valued  design parameters, avoiding the problematic optimization  by reducing it to a finite number of comparisons. Another attempt \cite{Kuck2006} proposes SMC to handle the optimisation part but restricts to static one-step design.
 In our work, we  consider both the continuous design and sequential setting, which  was referred to as an open question by \citet{Ryan2016review}.
  Recent ideas  using stochastic optimization  \cite{Blau2022,Kleinegesse2021,Foster2020,Foster2021,Foster2019,Huan2014} have since then paved the way to efficient optimisation formulation of sequential BOED but without consideration of
  SMC techniques and somewhat neglecting posterior distribution inference.
  A solution explored by \citet{Foster2019,Foster2020} uses mean field variational approximations of the posterior distributions but with no guarantee on their quality,
  %This has two drawbacks, at each step, a new approximation has to be computed from scratch and its quality is not guaranteed.
  while \citet{Foster2021,Blau2022} bypass the need for such posterior estimation using a reinforcement learning formulation.
  In this work, we show that superior performance than these previous solutions can be obtained with SMC, from which we also derive new good features. We provide theoretical results and guarantees that have not been formulated before in {\it modern} BOED and we shed a new light on tempering by showing that it can be crucial in SMC-based sequential BOED, to face the contradiction between searching for the largest information gain and maintaining good sampling performance.

\section{Sequential Bayesian Optimal Experimental Design (BOED)}

The Bayesian framework is a unified way to account for prior information via a probability distribution $p(\thetab)$, for
uncertainties about the observations $\yv$ through a distribution $p(\yv | \thetab, \xib)$, and for a design criterion (also called utility function) $F(\xib,\thetab, \yv)$ describing the experimental aims. The prior is assumed
to be independent of $\xib$ and  $p(\yv | \thetab, \xib)$ available in closed-form.

\paragraph{Expected Information Gain (EIG).}
There exists various utility functions $F$ depending on the targeted task \cite{Ryan2016review,Kleinegesse2021}. In this work, we focus on parameter estimation and consider an information-based utility leading to the so-called  expected information gain (EIG). The EIG, denoted by $I$, admits several equivalent expressions, see {\it e.g.} \cite{Foster2019}. It can be written as the expected loss in entropy when accounting for an observation $\yv$ at $\xib$ (eq.~(\ref{def:I})) or as a mutual information or expected Kullback-Leibler divergence (eq. (\ref{def:EIGpost})). Using $p(\yv,\thetab | \xib)=p(\thetab | \yv, \xib) p(\yv| \xib)= p(\yv | \thetab, \xib) p(\thetab)$,
\begin{align}
I(\xib) & =  \Exp_{p(\yv | \xib)}[H(p(\thetab)) - H(p(\thetab | \Yv, \xib)]  \label{def:I} \\
&=  \Exp_{p(\yv | \xib)}\left[\text{KL}(p(\thetab | \Yv, \xib), p(\thetab))\right] \label{def:EIGpost}
\end{align}
where random variables are indicated with uppercase letters, $\Exp_p[\cdot]$ denotes the expectation with respect to $p$, KL the Kullback-Leibler divergence and  $H(p(\thetab))= -\Exp_{p(\thetab)}[\log p(\thetab)]$
 is the entropy of $p$.
We thus look for $\xib^*$ satisfying
\begin{equation}
\xib^* \in \arg\max_{\xib \in \Rset^d} I(\xib)\;. \label{def:xistar}
\end{equation}
Before optimizing $I(\xib)$, evaluating $I(\xib)$ is often difficult due to the intractability of $p(\thetab | \yv, \xib)$ and $p(\yv | \xib)$.

\paragraph{Sequential design.}
Solving (\ref{def:xistar}) is a {\it static}  or {\it one-step} design problem. A single $\xib$ or multiple $\{\xib_1, \cdot, \xib_K\}$ are selected prior to any observation, measurements $\{\yv_1, \cdot, \yv_K\}$ are made for these design parameters and the experiment is stopped. The prior $p(\thetab)$ can be used to encode previous observations but in static design, the selected designs depend only on the model.
 In contrast, in sequential or iterated design, $K$ experiments are planned sequentially to construct an adaptive strategy, meaning that for the k$^{th}$ experiment, the best $\xib_k$ is selected taking into account the previous design parameters and associated observations $\Db_{k-1} =\{(\yv_1,\xib_1), \cdot , (\yv_{k-1}, \xib_{k-1})\}$.  Then, $\yv_k$ is measured at $\xib_k$ and $\Db_{k}$ is updated into $\Db_k = \Db_{k-1} \cup (\yv_k, \xib_k)$.
 This approach is referred to as {\it greedy} or {\it myopic}, in that the next design $\xib_k$ is chosen as if it was the last one. Non-myopic approaches exist, using for instance reinforcement learning principles \cite{Blau2022,Foster2021} but with another layer of complexity and performance that are not always superior, see  \cite{Blau2022} or our results in Section \ref{sec:exp}.
In this work, we limit ourselves to the greedy approach, replacing in (\ref{def:I}) or (\ref{def:EIGpost}) the prior $p(\thetab)$ by our current belief on $\thetab$, namely $p(\thetab | \Db_{k-1})  = p(\thetab | \yv_1, \xib_1, \cdot ,  \yv_{k-1},  \xib_{k-1})$, and to solve iteratively for
\begin{align}
\xib_k^* &\in \arg\max_{\xib \in \Rset^d} I_k(\xib), \label{def:seqd}
\end{align}
\vspace{-.5cm}
\begin{align}
\mbox{where } & I_k(\xib)  =  \Exp_{}[H\!(p(\thetab | \Db_{k-1}))\!- \!H\!(p(\thetab  | \Yv, \xib, \!\Db_{k-1}))] \nonumber \\
  &=  \Exp_{}\left[\text{KL}(p(\thetab  | \Yv, \xib, \!\Db_{k-1}), p(\thetab | \Db_{k-1}))\right]
    \label{def:Ik}
    \end{align}
    and $\Exp$  is  with respect to $p(\yv | \xib, \Db_{k-1})$.
Observations are   assumed conditionally independent so that
$\!p(\thetab | \Db_k) \!\propto \!p(\thetab) \prod_{i=1}^k\! p(\yv_i | \thetab, \xib_i)\!$ which also leads to
\begin{align}
p(\thetab | \Db_k) \propto p(\thetab | \Db_{k-1}) \; p(\yv_k | \thetab, \xib_k) \; . \label{eq:ci}
\end{align}

\paragraph{EIG contrastive bound optimization.}
Going back to the optimization in (\ref{def:xistar}),  we focus  on the continuous design case and assume  that quantities are differentiable when needed and that gradients are well defined. This is generally not a restrictive assumption except for discrete design spaces that may require specific treatments, see {\it e.g.} \cite{Blau2022,Drovandi2014}. A standard gradient ascent algorithm would consist, at iteration $t$, of updating
 $ \xib_{t+1}  =    \xib_t + \gamma_t \; \nabla_\xib{I}(\xib)_{| \xib =\xib_t}$ with a stepsize $\gamma_t$, but in practice both $I(\xib)$ and its gradient $ \nabla_\xi{I}(\xib)$  are intractable.
However, they can both be expressed as expectations, which naturally leads to consider stochastic approximation approaches \cite{Borkar2008},
among which the most popular is the Stochastic Gradient (SG) algorithm. In a BOED setting, if $\nabla_\xib{I}(\xib)$ is expressed as an expectation  $\nabla_\xib I(\xib) =\Exp[f(\xib, \Xv)]$ over a random variable $\Xv$,
the SG iteration writes $\xib_{t+1}  =    \xib_t + \gamma_t \; f(\xib_t, \xv_t)$ with $\xv_t$ a realisation of $\Xv$.
This assumes that we can differentiate under the integral sign in (\ref{def:EIGpost}).
There exist different ways to differentiate, including the popular  reparametrization trick, but SG for $I(\xib)$ remains difficult to perform due to the intractability of the integrand in (\ref{def:EIGpost}). An alternative  has been proposed by \citet{Foster2019,Foster2020} referred to as a variational approach. It consists of optimizing a tractable lower bound  of $I(\xib)$ and computing $\xib^*$ via an alternate maximization.
 We also consider such a bound, $I_{PCE}$, introduced by \citet{Foster2020} and  named the Prior Contrastive Estimation (PCE) bound. It is based on
 contrastive samples from  $L$ additional variables $\thetab_\ell$, for $\ell=1:L$, distributed following the prior $p(\thetab)$ {as $\thetab$, which is} rewritten as $\thetab_0$. $I_{PCE}$ is defined as
 \begin{equation}
I_{PCE}(\xib) = \Exp_{p(y| \thetab_0, \xib)\prod_{\ell=0}^L p(\thetab_\ell) }\left[F(\xib, \thetab_0, \cdot, \thetab_L, \Yv)\right] \label{def:ipce}
\end{equation}
  $$\mbox{with } F(\xib, \thetab_0, \cdot, \thetab_L, \yv)\! = \!\log \frac{p(\yv | \thetab_0, \xib)}{\frac{1}{L+1} \sum_{\ell=0}^L p(\yv | \thetab_\ell, \xib)}.$$
$I_{PCE}$ is a lower bound  $I(\xib) \geq  I_{PCE}(\xib) $  and the bound is tight when $L$ tends to $\infty$ (see \citet{Foster2020} for a proof). It is tractable as all expressions $p(\yv | \thetab_\ell, \xib)$ are tractable, and its gradient requires only the gradient $\nabla_\xib p(\yv | \thetab, \xib)$. Stochastic approximation can be applied to maximize $I_{PCE}$ as $\nabla_\xib I_{PCE}(\xib)$ can be expressed as an expectation via reparametrization. More specifically,  we assume that there exists a transformation $T^{\xib}_{\thetab_0}$ such that $\Yv=T^{\xib}_{\thetab_0}(U)$ with $U \in {\cal U}$ independent of $\xib$ and $\thetab_0$ and easy to simulate, {\it e.g.} $U$ is a standard Gaussian. It follows under some mild conditions specified in the Appendix that
\begin{align}
\!\! \nabla_{\!\xib}\!I_{\!PCE}(\xib) \!=  \! \Exp_{p(u)\!\prod_{\ell\!=\!0}^L \!p(\thetab_\ell)}\!\!\left[\!\nabla_{\!\xib}\! F(\xib, \! \thetab_0,  \cdot, \!\thetab_L,\! T^{\xib}_{\thetab_0}\!(U))\!\right]  \label{def:gradipce}
 \end{align}
A similar bound  and its gradient can be derived for sequential design optimization (\ref{def:seqd}).
 Using conditional independence (\ref{eq:ci}), at each step $k$,  $p(\thetab)$  only needs to be replaced  by  the current posterior
$p(\thetab | \Db_{k-1})$,
{
\begin{align}
\!\!I^k_{PCE}(\xib)\! &= \nonumber\\
&\!\Exp_{p(\yv|\thetab_0,\xib))\!\prod_{\ell=0}^L p(\thetab_\ell| \Db_{k-1}\!)}\!\left[F(\xib, \thetab_0, \cdot, \thetab_L,\! \Yv)\right] \!\! \label{def:ipcek}
\end{align}
\vspace{-.5cm}
\begin{align*}
\!\! \nabla_{\!\!\xib}\!I^k_{\!PCE}(\xib)\! &= \!  \Exp_{p(u)\!\prod_{\ell=0}^L\! p(\thetab_\ell| \Db_{k-1})}\!\!\left[\! \nabla_{\!\!\xib} \!F\!(\xib, \! \thetab_0,  \cdot, \!\thetab_L,\! T^{\xib}_{\thetab_0}\!(U)\!)\!\right] . %\label{def:gradipcek}
 \end{align*}
 }%
The stochastic gradient  algorithm  is summarized  in Algorithm \ref{alg:sgd}, with the additional possibility to estimate gradients with minibatches of size $N_t$, in line 5.

%\LinesNumbered
\begin{algorithm}[h!]
	\caption{SG with minibatches $\!(N_t)_{t=1:T}$  at step $k$ to optimize $I^k_{PCE}$ in (\ref{def:ipcek}) }\label{alg:sgd}
    \begin{algorithmic}[1]
   \State Set $T$ iterations, $\xib_0$, stepsizes $(\gamma_t)_{t=1:T}$
	\While{$t \leq T$}
		\State Sample  $\thetab^i_{\ell, t} \! \sim p(\thetab | \Db_{k-1})$,  $\ell\!= \!0\! :\! L, i\!=\!1\!: \!N_t$
		\State Sample $u^i_{t} \sim  p(u),  \quad \mbox{ for $i=1: N_t$}$
   \State Set $\!\nabla_{\!t+1} \!\!= \!\!\frac{1}{N_t} \!\!\sum\limits_{i=1}^{N_t} \!\!\nabla_\xib \!F(\xib,\!  \thetab^i_{0,t},  \cdot ,\! \thetab^i_{L,t}, \!T^{\xib}_{\thetab^i_{0,t}}\!\!(u^i_{t})\!)|_{\xib = \xib_t}\!\!$
    \State Update $ \xib_{t+1} = \xib_t + \gamma_t  {\nabla}_{t+1}$
    \EndWhile
	\State \Return  $\xib_k^*=\xib_T$ or a Polyak averaging value
        \end{algorithmic}
\end{algorithm}

    	%%\begin{algorithmic}[1]
%   Set $T$ iterations, $\xib_0$, stepsizes $(\gamma_t)_{t=1:T}$ \\
%		 \While{$t \leq T$}{
%				    Sample  $\thetab^i_{\ell, t} \! \sim p(\thetab | \Db_{k-1})$,  $\ell\!= \!0\! :\! L, i\!=\!1\!: \!N_t$  \\
%				    Sample $u^i_{t} \sim  p(u),  \quad \mbox{ for $i=1: N_t$}$  \\
%   Set $\!\nabla_{\!t+1} \!\!= \!\!\frac{1}{N_t} \!\!\sum\limits_{i=1}^{N_t} \!\!\nabla_\xib \!F(\xib,\!  \thetab^i_{0,t},  \cdot ,\! \thetab^i_{L,t}, \!T^{\xib}_{\thetab^i_{0,t}}\!\!(u^i_{t})\!)|_{\xib = \xib_t}\!\!$ \\
% Update $ \xib_{t+1} = \xib_t + \gamma_t  {\nabla}_{t+1}$}
%		\Return  $\xib_k^*=\xib_T$ or a Polyak averaging value
        %%\end{algorithmic}

Optimizing the contrastive bound $I^k_{PCE}$ requires then sampling $\thetab_0, \cdot, \thetab_L$ from $p(\thetab | \Db_{k-1})$ (Algorithm \ref{alg:sgd} line 3). This is more costly than sampling from the prior, as $p(\thetab | \Db_{k-1})$ is  only know up to a normalizing constant (\ref{eq:ci}),
but can be efficiently dealt with using a SMC approach as detailed in the next section.

\section{Tempered Sequential Monte Carlo}
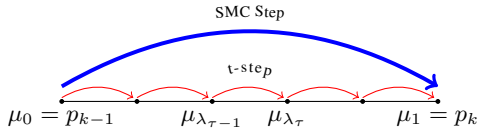
\begin{figure}[h!]
\centering
		\begin{tikzpicture}[scale=0.5]
		  \draw[-] (0,0) -- (10,0) node[right] {};

		  \foreach \x in {0, 2, 4, 6, 8, 10}
		    \filldraw[black] (\x,0) circle (0.05);

		  \draw[->, bend left=30, line width=1.5pt, blue, postaction={decorate,decoration={text along path,text align=center,text={|\tiny|SMC Step},raise=0.2cm}}] (0,0.4) to (10,0.4);
		  \draw[->, bend left=30, red] (0,0.1) to (1.9,0.1);
		  \draw[->, bend left=30, red] (2,0.1) to (3.9,0.1);
		  \draw[->, bend left=30, red, postaction={decorate,decoration={text along path,text align=center,text={|\tiny| t-step},raise=0.2cm}}] (4,0.1) to (5.9,0.1);
		  \draw[->, bend left=30, red] (6,0.1) to (7.9,0.1);
		  \draw[->, bend left=30, red] (8,0.1) to (9.9,0.1);

		  \node[below] at (4,0) {\footnotesize$\mu_{\lambda_{\tau-1}}$};
		  \node[below] at (6,0) {\footnotesize $\mu_{\lambda_\tau}$};
		  \node[below] at (0,0) {\footnotesize $\mu_{0} = p_{k-1}$};
		  \node[below] at (10,0) {\footnotesize $\mu_1 = p_k$};
		\end{tikzpicture}
  \caption{\footnotesize Tempered SMC, a SMC step from $p_{k-1}$ to $p_k$ (blue) performed in  ${\cal T}$ intermediate tempering steps (red).}
    \label{fig:tempering}
	\end{figure}

%\LinesNumbered
\begin{algorithm}
	\caption{Adaptive tempered SMC at step~$k$}\label{alg:SMC}
    \begin{algorithmic}[1]
		\State{Set $\tau\!=\!0, \lambda_0\!=\!0, M, ESS_{min} $, $M_\lambda$, resample}
		\State {Sample $\thetab_0^{1:M} \sim p(\thetab | \Db_{k-1}) = \mu_{\lambda_0}(\thetab)$}
		\State{Set $w_0^i\! =\! \frac{1}{M}$, $i\!=\!1:M$ and $\wv^{1:M}_0\! = \!\{w_0^1, \cdot, w_0^M\}$}
		\State{\textbf{while} $\lambda_\tau< 1$ \textbf{do}}
            \begin{tcolorbox}[colback=orange!10,colframe=orange!50!black, boxrule=0.01pt,boxsep=0.01pt, left=0.01pt, right=0.01pt, top=1pt, bottom=1pt, title style={halign=flush right,before upper=\strut}, width=0.94\columnwidth]
			\State{\hspace{0.6em} Set $\tau=\tau+1$}
			  \State{\hspace{0.6em} Set $\tilde{\thetab}_{\tau-1}^{1:M} \!=\!\textrm{resample}(\thetab^{1:M}_{\tau-1}, \wv^{1:M}_{\tau-1}))$~$\sim \mu_{\lambda_{\tau-1}}$}
			\State{\hspace{0.6em} Sample $\thetab^i_\tau \!\sim \!M_{\lambda_{\tau-1}}\!(\tilde{\thetab}_{\tau-1}^i, \cdot)$ for $i{=}1{:}M$,
            \State{\hspace{0.6em} Set $\thetab^{1:M}_\tau = \{\thetab_\tau^1, \cdot, \thetab_\tau^M\}$\quad
            \tikz[baseline=(text.base)]{\node[draw=orange, anchor=west, inner sep=2pt, font=\scriptsize, text=orange] (text) {Resampling and Markov kernel step};}}}
            \end{tcolorbox}
            \begin{tcolorbox}[colback=magenta!10,colframe=magenta!50!black, boxrule=0.01pt,boxsep=0.01pt, left=0.01pt, right=0.01pt, top=1pt, bottom=1pt,title style={halign=flush right,before upper=\strut}, width=0.94\columnwidth]
        	\State{\hspace{0.6em} Solve for $\gamma$,  $\frac{\left(\sum_{i=1}^M p\left(\yv_k | \thetab_\tau^i, \xib_k\right)^\gamma\right)^2}{\sum_{i=1}^M p\left(\yv_k |  \thetab_\tau^i,\xib_k\right)^{2\gamma}} = ESS_{min}$}
        	\State{\hspace{0.6em} Set $\lambda_\tau = \lambda_{\tau-1} + \gamma $
            \qquad \qquad \qquad \qquad \quad \tikz[baseline=(text.base)]{\node[draw=magenta, anchor=west, inner sep=2pt, font=\scriptsize, text=magenta] (text) {Tempered step};}}
            \end{tcolorbox}
            \begin{tcolorbox}[colback=cyan!10,colframe=cyan!50!black, boxrule=0.01pt,boxsep=0.1pt, left=0.01pt, right=0.01pt, top=1pt, bottom=1pt, title style={halign=right upper=\strut}, width=0.94\columnwidth]
		      \State{\hspace{0.6em} Set $\tilde{w}^i_\tau=p(\yv_k|\thetab^i_\tau,\xib_k)^\gamma , \,\displaystyle w^i_\tau=\frac{\tilde{w}^i_\tau}{\sum\limits_{j=1}^M\tilde{w}^j_\tau}$} \, \textrm{for}\,{i{=}1{:}M}
            \State{\hspace{0.6em} Set $\wv^{1:M}_\tau=\{w_\tau^1,\cdot\cdot\cdot,w_\tau^M\}$ \quad \qquad  \tikz[baseline=(text.base)]{\node[draw=cyan, anchor=east, inner sep=2pt, font=\scriptsize, text=cyan] (text) {Compute new weights};}}
        \end{tcolorbox}
     %\EndWhile
     \State \Return{$\thetab^{1:M}_k\!\!=\thetab_\tau^{1:M}\!\!, \wv_k^{1:M}\!\!= \wv_\tau^{1:M}$ for a particle approximation
      $p^M_k \!= \!\sum\limits_{i=1}^M\! w^i_k \delta_{\thetab^i_k}\!$
      of $p(\cdot| \Db_{k})=\mu_1$}
        \end{algorithmic}

\end{algorithm}
SMC  samplers  extend the idea of importance sampling by re-using samples  from one distribution to another, and  benefit from numerous theoretical results \cite{del2006sequential,Doucet2018,Naesseth2019,chopin2020introduction,dai2022invitation}.
More specifically, denoting ${\cal P}(\Thetab)$ the set of probability measures on $\Thetab$, the goal is to provide samples (called particles) from a sequence of probability distributions $\{p_k\}_{k=1:K}$ in ${\cal P}(\Thetab)$. To simplify, we deal with probability densities assuming absolute continuity with respect to the Lebesgue measure but the setting is more general, {\it e.g.}~\cite{chopin2020introduction}. A standard MCMC approach would require to build an ergodic kernel $M_k$ and to run a Markov chain from scratch for each $p_k$. In contrast, SMC samplers
provide  the possibility
 to approximate $p_{k}$ recycling samples from $p_{k-1}$.
 SMC samplers aim at propagating $M$  particles $\thetab^{1:M}_{k-1} = \{\thetab^1_{k-1},\cdot,\thetab^M_{k-1}\}$ and their corresponding weights $\wv^{1:M}_{k-1} = \{w^1_{k-1},\cdot,w^M_{k-1}\}$ in such a way that the empirical distribution $p^M_k$ of the particles at time $k$ converges to $p_k$ in some sense: meaning that for all integrable functions $\phi$,
$\Exp_{p^M_k}[\phi(\thetab)] = \sum_{i=1}^M w^i_k \phi(\thetab^i_k) \xrightarrow[M\rightarrow \infty]{}  \Exp_{p_k}[\phi(\thetab)].$
However, as showed by \citet{agapiou2017importance}, the number of particles $M$ required for an accurate {\it particle approximation} $p^M_k$ scales exponentially with the Kullback-Leibler distance between
the proposal $p_{k-1}$ and target $p_k$ distributions.
For $p_{k-1}(\thetab)=p(\thetab|\Db_{k-1})$ and  $p_k(\thetab)=p(\thetab |\Db_{k})$, this is problematic as EIG optimization aims at increasing this distance, see
 (\ref{def:seqd},\ref{def:Ik}).  Moving from $p(\thetab | \Db_{k-1})$ to $p(\thetab | \Db_{k})$ with just one SMC step  might  then yield poor results.
A solution is to consider
 {\it tempering}
 %\cite{chopin2020introduction,Naesseth2019,Syed2019}
 \cite{Neal2001,del2006sequential,Syed2019}
 to move along a sequence of probability distributions interpolating between $p(\thetab | \Db_{k-1})$ and $p(\thetab | \Db_{k})$. A tempering path is a sequence  of the form $\mu_{\lambda_\tau}$ with $0= \lambda_0 < \lambda_\tau < \ldots < \lambda_{\cal{T}} =1$ where  $\mu_{0} = p_{k-1}$ and $\mu_1 = p_k$ (Figure \ref{fig:tempering}).  Usually, only the initial and final distributions are imposed. Intermediate distributions $\mu_{\lambda_\tau}$  are not of interest so that the $\lambda_\tau$'s can be chosen as desired. A popular approach is to use what is know as the geometric path:
 $\mu_{\lambda_\tau}(\thetab) \propto  p_{k-1}(\thetab)^{1-\lambda_\tau} \;  p_{k}(\thetab)^{\lambda_\tau},$ which using (\ref{eq:ci}),  takes the form
  $\mu_{\lambda_\tau}(\thetab) \propto  p(\thetab | \Db_{k-1}) \;  p(\yv_k | \thetab, \xib_k)^{\lambda_\tau} $ or equivalently
	$\mu_{\lambda_\tau}(\thetab) \propto \mu_{\lambda_{\tau-1}}(\thetab) \; p(\yv_k | \thetab, \xib_k)^{\lambda_\tau-\lambda_{\tau-1}}.$
As setting the sequence $\lambda_\tau$ manually can be a challenging task with disappointing results, we follow the adaptive strategy proposed by \citet{jasra2011inference}. Given a user-set threshold $ESS_{min}$ interpreted as an effective sample size, at iteration $\tau$ of the tempered SMC procedure, given a current set of particles $\thetab_\tau^{1:M}$, we set recursively $\lambda_\tau= \lambda_{\tau-1}+\gamma$ with $\gamma$ the solution in $\left[0, 1 - \lambda_{\tau-1}\right]$ of the  equation
	$\frac{\left(\sum_{i=1}^M p\left(\yv_k | \thetab_\tau^i, \xib_k\right)^\gamma \right)^2}{\sum_{i=1}^M p\left(\yv_k |  \thetab_\tau^i,\xib_k\right)^{2\gamma}} = ESS_{min} \; .$ If $\gamma$ is not in $[0, 1 - \lambda_{\tau-1}]$, $\lambda_\tau$ is set to 1 and the tempering stops.
This is a relatively simple task to solve with numerical root finding. This procedure guarantees
that the SMC approximation error  remains stable over iterations, see \cite{jasra2011inference} for details. It can be interpreted as a way to control the Chi-square pseudo-distance between the successive distributions, see \cite{chopin2020introduction} Proposition 17.2.
Tempered SMC then requires like SMC an unbiased resampling scheme denoted by $resample(\thetab^{1:M},\wv^{1:M})$. Resampling is the action of drawing randomly from a weighted sample, so as to obtain an unweighted sample. Several unbiased resampling schemes are listed in Chapter 9 of \citet{chopin2020introduction} and studied in {\it e.g.} \cite{Gerber2019,Crisan2002}. The most standard one is {\it multinomial} resampling which draws samples independently according to their weights. Tempered SMC also requires a family of Markov kernels $(M_\lambda)_\lambda$ so that  $M_\lambda: \Thetab \rightarrow {\cal P}(\Thetab)$  leaves $\mu_\lambda$ invariant. Tempering is illustrated in Figure \ref{fig:tempering} and
in Algorithm \ref{alg:SMC}.

\section{Particle EIG contrastive bound approximation and optimization}
\label{Sec:combi}

Algorithms \ref{alg:sgd} and \ref{alg:SMC} can then be combined iteratively. We call the resulting algorithm PASOA. Algorithm \ref{alg:SMC} at step $k$  provides a particle approximation of $p(\thetab | \Db_k)$ used in Algorithm \ref{alg:sgd} line 3
at step $k+1$ to optimize the next EIG contrastive bound $I_{PCE}^{k+1}$ and get $\xib^*_{k+1}$. A new $\yv_{k+1}$ is then measured at $\xib^*_{k+1}$, $\Db_{k+1}$ is set to $\Db_k \cup \{\yv_{k+1},
\xib^*_{k+1}\}$ and Algorithm \ref{alg:SMC} is used again to get
a particle approximation of $p(\thetab | \Db_{k+1})$ etc.
More specifically, the $M$ weighted particles produced by Algorithm \ref{alg:SMC} are used to approximate the intractable $p(\thetab | \Db_k)$ and to simplify the sampling of the $L+1$ contrastive variables in line 3 of Algorithm \ref{alg:sgd} at step $k+1$. In practice, line 3 can be performed in different ways. We propose the following one, which has good  numerical and asymptotic properties (see Section \ref{sec:theo}).
Due to the use of generally large numbers $L$ (typically $L=200$) of contrastive samples, expectations of interest, {\it e.g.} (\ref{def:ipce},\ref{def:gradipce}),  are computed in  a large dimensional space $\Thetab^{L+1}$ even if $\Thetab$ is of moderate dimension. A simple way to mitigate the dimension impact is to start from $M= N(L+1)$ particles using a SMC procedure on $\Thetab$ and
partition them into $L+1$ disjoint subsets of $N$ particles, denoted by $\thetab_{k,\ell}^{1:N}$, for $\ell\!=\!0\!:\!L$, with their associated weights denoted by $W_{k,\ell}^{1:N}$ with $\sum_{i=1}^N  W_{k,\ell}^{i} =1$.
Seeing particles and  weights   as random variables,
if the $resampling$ procedure used in Algorithm \ref{alg:SMC} is so that the resampled values are independent conditionally on the previous particles, then the $L+1$ collections of random variables denoted by $\zeta_{k,\ell}^{N} = \{W_{k,\ell}^{i},\thetab^{i}_{k,\ell}\}_{i=1:N}$ are independent and identically distributed ({\it i.i.d.}) conditionally on the previous particles.
For instance, multinomial resampling  preserves conditional independence.  From the weighted particles produced by Algorithm \ref{alg:SMC}, we then derive  $L+1$
{\it i.i.d.}  random probability measures  $P_{k,\ell}^N =\sum_{i=1}^N  W_{k,\ell}^{i} \delta_{\thetab_{k,\ell}^i}$ ($\delta_\thetab$ is the Dirac measure at $\thetab$) and
 the gradient estimate  at each iteration $t$, Algorithm \ref{alg:sgd} line 5, is computed with $\theta_{0,t}^i, \ldots, \theta_{L,t}^i$ values  sampled independently from realizations $p_{k,\ell}^N$ of the respective independent particle approximations $P_{k,\ell}^N$ for $i=1:N_t$ and $\ell=0:L$.
We thus  define $P_{k,\ell}^N$ on $\Thetab$ and use in our SG algorithm  approximated distributions on $\Thetab^{L+1}$ which are  product measures $\otimes_{\ell=0}^L P_{k,\ell}^N$.
 It follows a
 SG procedure, detailed  in the  Appendix
 (Algorithm \ref{alg:sgdsmc}),
 which produces an estimator $\xib^*_{k+1,N}$  from  the optimization of  a particle approximation of $I^{k+1}_{PCE}$ defined as
\begin{align}
\!\!I^{k+1,N}_{\!PCE}\!(\xib)\! &= \! \Exp_{p(u)\!\prod_{\ell=0}^L {p}^N_{k,\ell}(\thetab_\ell)}\!\!\left[\! F\!(\xib, \! \thetab_0,  \cdot, \!\thetab_L,\! T^{\xib}_{\thetab_0}\!(U)\!)\!\right] \label{def:ipcekN}
\end{align}
The advantage of such a product form particle approximation over standard SMC lies in the following observation.
%With the use of contrastive samples, the quantities we target, e.g. equation (\ref{8}),involve $L + 1$ copies $(\thetab_0, \ldots,\thetab_L)$ and are on the product space $\Thetab^{L+1}$, which is much higher dimensional than$\Thetab$ when $L$ is large.
Standard SMC would consist in using $ \prod_{\ell=0}^L\! p(\thetab_\ell) \approx P_k^N(\thetab_0, \cdot, \thetab_L)$ where
\begin{align}
P_k^N &=
    \sum_{i=1}^N \left( \prod_{\ell=0}^L W_{k,\ell}^i\right)\! \delta_{(\thetab^i_{k,0}, \cdot, \thetab^i_{k,L})} \label{eq:pf1} \; .
\end{align}
As an alternative, with the same $M$ weights and particles, our {\it product form} approach is using $ \prod_{\ell=0}^L p(\thetab_\ell) \approx  \prod_{\ell=0}^L p^N_{k,\ell}(\thetab_\ell)$,
%where $p^N_{k,\ell}(\thetab_\ell) \propto \sum_{i=1}^N W_{k,\ell}^i \; \delta_{\thetab_{k,\ell}^i}$. The former
which can be rewritten as a sum, {\it i.e.}
\begin{align}
    \prod_{\ell=0}^L p^N_{k,\ell} & = \sum_{i_0=1}^N  \cdot  \sum_{i_L=1}^N \left( \prod_{\ell=0}^L W_{k,\ell}^{i_\ell}\right) \delta_{(\thetab^{i_0}_{k,0}, \cdot, \thetab^{i_L}_{k,L})}\; . \label{eq:pf2}
\end{align}
There are two main differences between (\ref{eq:pf1}) and (\ref{eq:pf2}). First, (\ref{eq:pf2}) is more statistically efficient: the
$\thetab_{k,\ell}^i$
are {\it i.i.d.}  and so all permutations $(\thetab^{i_0}_{k,0}, \cdot, \thetab^{i_L}_{k,L})$
of these samples, for $1\leq i_0, \ldots, i_L \leq N$, are identically distributed. Hence (\ref{eq:pf2}) averages over $N^{L+1}$
tuples while its conventional counterpart (\ref{eq:pf1}) only averages over $N$ tuples. This increase in tuple number leads to a decrease in estimators variance. With the same particles, product-form
(\ref{eq:pf2}) makes the most out of every sample available.
Second, in contrast to tuples in (\ref{eq:pf1}), those in (\ref{eq:pf2}) are not independent because the same $\thetab_{k,\ell}^i$  appears in many terms of the sum. As a consequence, using product form approximations requires  to generalize standard SMC results, which rely on the independence of the terms in the sum.
 In the next section, we take that into account to show  the convergence of the quantities involved in PASOA.

\section{Asymptotic properties}
\label{sec:theo}

With the contrastive bound approach, to increase approximations quality, we can  play with $L$ the number of contrastive samples and $N$ the number of particles. It has already been shown by \citet{Foster2020} that increasing $L$  improves the quality of the $I_{PCE}$ bound that becomes tight, {\it i.e.} to the limit equal to the EIG. We then study the behavior of our approximations when the number $N$ of particles tends to $\infty$.
The sequential design  values  produced can be seen as realizations of random estimators $\{\xib_{k+1,N}^*\}_{N\geq 1}$ targeting points of maximum of random criterion functions $\{I_{PCE}^{k+1,N}(\xib)\}_{N\geq 1}$. We use here the same notation for random quantities and their realizations. We first establish the convergence of the criterion functions  adapting standard SMC techniques in Proposition \ref{propL2}, and then the convergence of their points of maximum, {\it i.e.} the consistency of the $\{\xib_{k+1,N}^*\}_{N\geq 1}$ to the optimal designs in Proposition \ref{prop:consTSMC}.

\paragraph{Convergence of product particle approximations.}

%Due to the use of generally large numbers $L$ (typically $L=200$)
With the use of contrastive samples, expectations
%of interest, {\it e.g.} (\ref{def:ipce},\ref{def:gradipce}),
are computed in
%a large dimensional space
$\Thetab^{L+1}$. Particles could then be considered as $(L+1)$-dimensional elements in this space so that standard SMC convergence results could apply \cite{chopin2020introduction}, Chapter 11. However, working in a $(L+1)$-dimensional space is more challenging for SMC and not necessary for the type of convergence result we need. As importance sampling, SMC suffers from the weights' variance which scales unfavorably
with the dimension of the problem \cite{Naesseth2019,chopin2020introduction}.
In contrast, in Section \ref{Sec:combi}, we define $P_{k,\ell}^N$ on $\Thetab$ and use in our SG algorithm (Algorithm \ref{alg:sgdsmc} in Appendix) approximated distributions on $\Thetab^{L+1}$ which are  product measures $\otimes_{\ell=0}^L P_{k,\ell}^N$.
Such products have been used in various settings but have been only recently studied in a more general way \cite{Kuntz2022}. However, theoretical results therein do not cover the use of product form estimators within SMC samplers. Other papers that consider {\it structured} SMC settings \cite{Rebeschini2015,Lindsten2017,Kuntz2023,Aitchison2019} differ from our work in key aspects  further discussed in Appendix Section \ref{sec:prodapp}.

Let ${\cal C}_b(\Thetab^{L+1})$
denote the set of functions $\phi: \Thetab^{L+1} \rightarrow \Rset$  that are measurable and bounded, and
let  $||\phi ||_\infty$ denote the supremum norm $||\phi ||_\infty = \sup_{\thetab \in \Thetab^{L+1}} |\phi(\thetab)|$.
Using the notation and terminology of \citet{chopin2020introduction} and previous authors before them, we introduce the potential function ${G}_{k,\tau}$ which in our setting is equal to ${G}_{k,\tau}(\thetab) = p(\yv_k | \thetab, \xib_k)^\gamma$ and used in Algorithm \ref{alg:SMC} to compute weights. We simplified the notation but $\gamma$ may also depend on $k$ and $\tau$. More specifically, in standard tempering, $\gamma$ is a preset constant but in Algorithm \ref{alg:SMC}, we use an adaptive tempering where $\gamma$ depends on previous particles. Although adaptive tempering schemes are often more challenging to study, {\it e.g.} \cite{Salomone2018,Beskos2016,Delmoral2012},
this is not problematic in our setting as $\gamma \leq 1$ and all we require is $G_{k,\tau}$ to be bounded.

For our result, the assumption that $p(\yv_k | \thetab, \xib_k)$ is bounded and strictly positive is enough  but more general results and cases are available in \cite{Beskos2016,jasra2011inference}.
The following proposition establishes the convergence in L$_2$-norm, which also implies the convergence in L$_1$-norm and in probability.

Following \citet{chopin2020introduction,Crisan2002}, we can establish similarly L$_p$-norm ($p>2$) and almost-sure convergence but  convergence in probability is sufficient for consistency of the design estimators.
We denote by $\zeta_{k,L}^{N}$ the collection of random variables  $\zeta_{k,L}^{N} = \{W_{k,\ell}^{i},\thetab^{i}_{k,\ell}\}_{i=1:N, \ell=0:L}$, and use the short notation $p_k^{\otimes L+1}[\phi]$ for $p_k^{\otimes L+1}[\phi] = \Exp_{\prod_{\ell=0}^L p(\thetab_\ell|\Db_k)}[\phi(\thetab_0, \cdot, \thetab_L)]$.

\begin{proposition}[\bf L$_2$ convergence] \label{propL2}
Using Algorithm \ref{alg:SMC} with multinomial resampling and assuming that all potential functions ${G}_{k,\tau}$ are upper bounded,   there exists a constant $c_k$ such that,  for all functions $\phi \in {\cal C}_b(\Thetab^{L+1})$,
\begin{align*}
    & \Exp_{\zeta_{k,L}^{N}}\!\!\left[\!\left\{ \!\sum\limits_{i_0=1}^N\!\cdot\!\sum\limits_{i_L=1}^N \prod_{\ell=0}^L W_{k,\ell}^{i_\ell} \; \phi(\thetab^{i_0}_{k,0}, \cdot, \thetab^{i_L}_{k,L})\!  - p_k^{\otimes L+1}[\phi] \!\right\}^{\!\!2} \!\right] \\
     &\leq c_k  \; c_{N,L} \; \Vert\phi \Vert_\infty^2
     \end{align*}
with $c_{N,L} = 1- (1 -\frac{1}{N})^{L+1} \approx \frac{L+1}{N}$ and where the expectation is taken over all the realizations of the random tempered SMC method, or equivalently on $\zeta_{k,L}^{N}$.
  \end{proposition}
The proof is detailed in Appendix and extends the proof in Chapter 11 of \citet{chopin2020introduction} which is itself  mainly based on \cite{Crisan2002}.
Note that the first term in the outer expectation above corresponds to the expectation of $\phi$ with respect to $\otimes_{\ell=0}^L P^N_{k,\ell}$, which is a random variable as the $P^N_{k,\ell}$ are random measures.
For any $\xib \in {\cal E}$, when $\phi$ is set to $\phi_\xib( \thetab_0, \cdot, \thetab_{L}) = f_{PCE}(\xib,\thetab_0, \cdot, \thetab_L)$, with
$f_{PCE}(\xib,\thetab_0, \cdot, \thetab_L)=   \Exp_{p(u)}\left[F(\xib, \! \thetab_0,  \cdot, \!\thetab_L,\! T^{\xib}_{\thetab_0}\!(U))\right] $  assumed to be bounded on $\Thetab^{L+1}$ for all $\xib$,  then
$p_k^{\otimes L+1}[\phi_\xib] =  I^{k+1}_{\!PCE}(\xib),$
the PCE lower bound (\ref{def:ipcek}) we seek to optimize at step $k+1$.
Similarly, the expectation of $\phi_\xib$ with respect to
$\otimes_{\ell=0}^L P^N_{k,\ell}$ is
$I^{k+1,N}_{\!PCE}(\xib)$ in (\ref{def:ipcekN}).
 It follows from Proposition \ref{propL2}
 that
 $$  \Exp_{\zeta_{k,L}^{N}}\!\left[\left\{{I}^{k+1,N}_{\!PCE}(\xib) - I^{k+1}_{\!PCE}(\xib)\right\}^2 \right]  \leq c_k \; c_{N,L} \;\Vert\phi \Vert_\infty^2,$$
or in other words
 that for all $\xib \in {\cal E}$, the sequence of random variables  $\{I^{k+1,N}_{\!PCE}(\xib)\}_{N \geq 1}$ converges in L$_2$-norm to $I^{k+1}_{\!PCE}(\xib)$ when $N$ tends to $\infty$.
 Using Chebyshev's inequality this also implies the convergence in probability, pointwise in $\xib$, that is for all $\xib$, for all $\epsilon>0$,
\begin{equation}
 \lim_{N\rightarrow \infty} p_ {\zeta_{k,L}^{N}}\!\left( \left|{I}^{k+1,N}_{\!PCE}(\xib) - I^{k+1}_{\!PCE}(\xib)\right| \geq \epsilon\right) = 0 .\label{eq:cvp}
 \end{equation}

To further comment this result and give more insight on the constants involved, we note that interestingly, if the normalized potential functions are bounded by 1 (see Appendix Section \ref{sec:conv} for details), the constant denoted by $c_k$ in Proposition \ref{propL2}  remains low and decreases when $L$ increases. Otherwise, the constant may increase very fast with $k$ and, even if $k$ is relatively low in BOED, the bound may become uninformative for finite $N$. This is actually a well referenced behavior in SMC, see {\it e.g.} \cite{chopin2020introduction}. In our work, we limited the presentation to {\it simple} conditions but with stronger assumptions, in particular on the Markov kernels, it is possible to limit the growing of $c_k$ over steps $k$, see Section 11.4. in \cite{chopin2020introduction}.

\paragraph{Consistency of the sequential design estimators.}
Under additional  assumptions, specified in Proposition \ref{prop:consTSMC}, it follows from Proposition \ref{propL2} that the design parameters found at each step of our procedure tend to the ideal ones.

Assume $I^{k+1}_{\!PCE}(\xib)$ reaches its maximum in $\xib^*_{k+1}$ and consider a sequence $\{ \xib^*_{k+1,N}\}_{N \geq 1}$ of estimators defined by
\begin{align} \xib^*_{k+1,N} &\in \arg\max_{\xib \in {\cal E}} I^{k+1,N}_{\!PCE}(\xib) \;. \label{def:argmax}
\end{align}
 Proposition \ref{prop:consTSMC} states that  $\{ \xib^*_{k+1,N}\}_{N \geq 1}$ are consistent estimators of $\xib^*_{k+1}$, {\it i.e.} that for all $\epsilon > 0$,
\begin{eqnarray}
 \lim_{N\rightarrow \infty} p_ {\zeta_{k,L}^{N}}\!\left( \Vert  \xib^*_{k+1,N} -  \xib^*_{k+1}\Vert \geq \epsilon\right) &=& 0 .\label{eq:cons}
 \end{eqnarray}
 The assumptions in Proposition \ref{prop:consTSMC} are in a simplified form.
 More general situations could be handled but would require more complex developments while not changing the general idea.
To simplify the presentation, our first assumption (A1) is that the design space ${\cal E}$ is  compact.  The second assumption (A2) is to ensure that
the point of maximum $\xib^*_{k+1}$ is isolated or well-separated and that only design values in the neighborhood of $\xib^*_{k+1}$ reaches values close to $ I^{k+1}_{\!PCE}(\xib^*_{k+1})$ (see Lemma \ref{lem:hypA2} in Appendix). Assumption (A3) indicates that estimators $\xib^*_{k+1,N}$ could be points of near maximum only. The formulation of  (A3) uses additional variables $\rho_N$ that  account for the fact that estimators $ \xib^*_{k+1,N}$ are usually only approximations of the maxima in (\ref{def:argmax}), as in practice the optimization task is solved numerically with a given precision that can be controlled in the sense that a bound on the approximation error can be provided.
If we assume that $\xib^*_{k+1,N}$ is an exact maximizer of ${I}^{k+1,N}_{\!PCE}(\xib)$, then (A3) is trivially satisfied with $\rho_N=0$. In our setting, $\xib^*_{k+1,N}$ is obtained by a stochastic gradient algorithm for which we could exhibit the $\rho_N$ sequence but this is a particular case left to the interested reader.
This point of view, the assumptions and the technique of proof we are using, are similar to the ones used for establishing asymptotic properties of $M$-estimators \cite{vaart_1998,Vaartwellner1996}. A  proof is given in Appendix, using the pointwise convergence  (\ref{eq:cvp}) from Proposition \ref{propL2} and several intermediate results.

\begin{proposition}[\bf Consistency] \label{prop:consTSMC}
Assume  %B1 B3 A3
\begin{assumption}
${\cal E} \in \Rset^d$ is a compact set.
    \end{assumption}
    \begin{assumption}
For all $\xib \not = \xib^*_{k+1}$, $I^{k+1}_{\!PCE}(\xib) < I^{k+1}_{\!PCE}(\xib^*_{k+1})$
    \end{assumption}
    \begin{assumption}There exists a sequence  $\{\rho_N\}_{N\geq 1}$ of positive random variables  and a sequence of random variables  $\{\xib^*_{k+1,N}\}_{N\geq 1}$ in ${\cal E}$ that satisfy
$$\forall \epsilon>0, \quad \lim\limits_{N\rightarrow \infty} p_ {\zeta_{k,L}^{N}}\left(\rho_N \geq \epsilon \right) =0$$
$$  \lim\!\inf\limits_{N\rightarrow \infty}\! p_{\zeta_{k,L}^{N}}\!\!\left( \!{I}^{k+1,N}_{\!PCE}\!(\xib^*_{k+1,N}) \!\geq \! I^{k+1,N}_{\!PCE}\!(\xib^*_{k+1}) -\rho_N \! \right)\! =\!1$$
    \end{assumption}
Then the sequence  $\{ \xib^*_{k+1,N}\}_{N\geq 1}$ is consistent, {\it i.e.} for all $\epsilon >0$,
$$  \lim\limits_{N\rightarrow \infty} p_ {\zeta_{k,L}^{N}}\left( \Vert \xib^*_{k+1,N} - \xib^*_{k+1}\Vert \geq \epsilon\right) =0. $$
\end{proposition}

\section{Numerical experiments}
\label{sec:exp}
To benchmark our method in terms of information gained, we use
the {\it sequential prior contrastive estimation} (SPCE) and {\it sequential nested Monte Carlo} (SNMC) bounds introduced in \cite{Foster2021} and used in \cite{Blau2022}. %The expression can be evaluated using only the design and observation sequences.
For a number $K$ of experiments leading to $\xib_1, \cdot, \xib_K$, and $L$ contrastive variables, SPCE and SNMC are respectively lower and upper bounds for
the total EIG, {\it i.e.} the expected information
gained from the entire sequence $\xib_1, \cdot, \xib_K$ and they become tight when $L$ tends to $\infty$.
Their exact expressions are given in Appendix Sections \ref{sec:spce} and \ref{sec:snmc}.
They have the advantage of using only samples from the prior $p(\thetab)$ and not from the successive posterior distributions, thus not considering posterior approximation errors and making them  convenient criteria to compare methods on design sequences only.
{Methods can be compared via their  [SPCE, SNMC] intervals which contain the total EIG.}
However, one benefit of our approach is to also provide accurate posterior estimation for subsequent accurate parameter estimation. To assess this aspect, we use the L$_2$ Wasserstein distance between the  weighted particles yielded by our method and the true parameter $\thetab$.
We consider two other  recent approaches, namely a reinforcement learning-based approach RL-BOED from \cite{Blau2022} and the {\it variational prior contrastive estimation} VPCE of \citet{Foster2020}.
RL-BOED is a non-myopic approach, while VPCE involves iteratively optimizing $I_{PCE}$ in a myopic manner but estimating posterior distributions with variational
inference.
We also compare with a non tempered version of our approach and with a random baseline, where the observations $\{\yv_1, \cdot , \yv_K\}$ are simulated with designs generated randomly.
For methods that do not provide  posterior estimations or poor quality ones (RL-BOED, VPCE, Random), we compute Wasserstein distances on posterior samples obtained by using tempered SMC on their design and observation sequences. In contrast, naive SMC Wasserstein distances are computed on the SMC posterior samples to better assess the impact of tempering via the comparison with PASOA.

%\vspace{-0.9cm}
\paragraph{Source location.}

We consider the 2D  location finding experiment used in \cite{Foster2021,Blau2022}.
It consists of $S$  hidden sources in $\Rset^2$ whose locations $\thetab=\{\thetab_1, \cdot , \thetab_S\}$ are unknown.
Each  source emits a signal  whose intensity attenuates according to the inverse-square law.  The measured signal is
the superposition of all these signals. The design problem is to choose where to make the measurements to best learn the source locations. If a measurement is performed at a point  $\xib \in \Rset^2$, the signal strength is
 $\mu(\thetab,\xib) = b + \sum_{s=1}^S \frac{\alpha_s}{m + ||\thetab_s - \xib||_2^2}$ where $\alpha_s, b$ and $m$ are constants.
A standard Gaussian prior  is assumed for each $\thetab_s \sim {\cal N}(0, \Ib)$ and
the likelihood is assumed
log-normal,  $(\log \yv  | \thetab, \xib)\sim {\cal N}(\log \mu(\thetab,\xib), \sigma)$ with standard deviation $\sigma$.
We set  $S=2$, $\alpha_1=\alpha_2 =1$, $m=10^{-4}, b=10^{-1}$, $\sigma=0.5$ and we plan $K=30$ successive design optimisations.
The Markov kernel is that of a Metropolis-Hasting scheme with a Gaussian proposal centered at the current particle with a variance set to the empirical variance of the particles.
We use $ L=200$ contrastive variables for the  $I_{PCE}^k$ bound. Algorithm \ref{alg:SMC} is used to get $N=100$ simulations $\thetab_\ell^{1:N}$ of each contrastive variable. The Adam algorithm \cite{kingma2014adam} is then used  with standard hyperparameters to perform the stochastic gradient.

\begin{figure}[h!]
 \centering
    \includegraphics[trim={.3cm 0.7cm .2cm 0},clip,  width=0.49\linewidth]{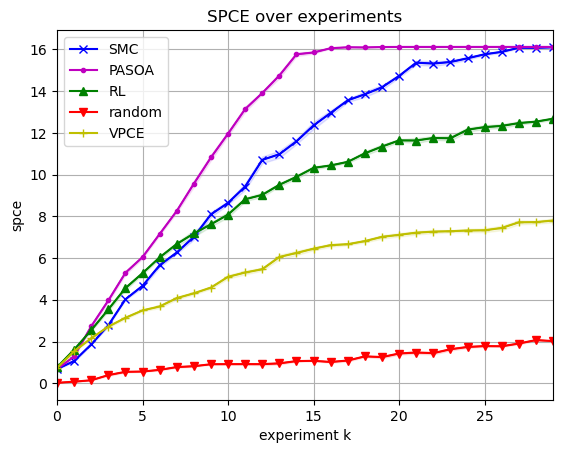}
    \includegraphics[trim={0.7cm 0.7cm 0.2cm 0},clip,width=.48\linewidth]{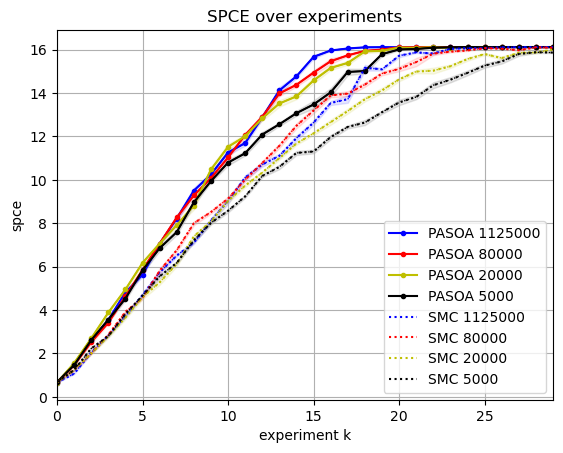}
\\
     \includegraphics[trim={.3cm 0.7cm 0.2cm 0},clip, height=3cm, width=0.49\linewidth]{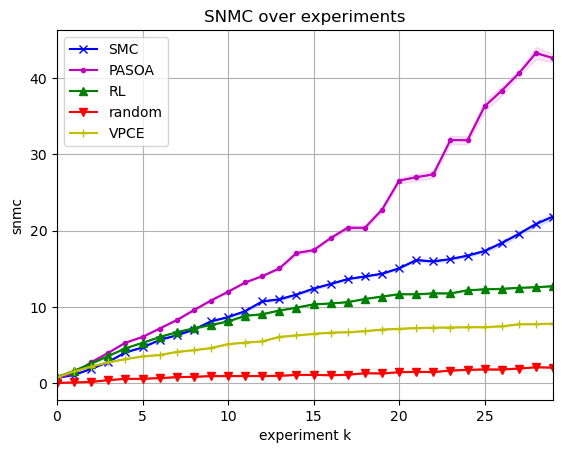}
     \includegraphics[trim={0.7cm 0.7cm 0.2cm 0},clip, height=3cm, width=0.49\linewidth]{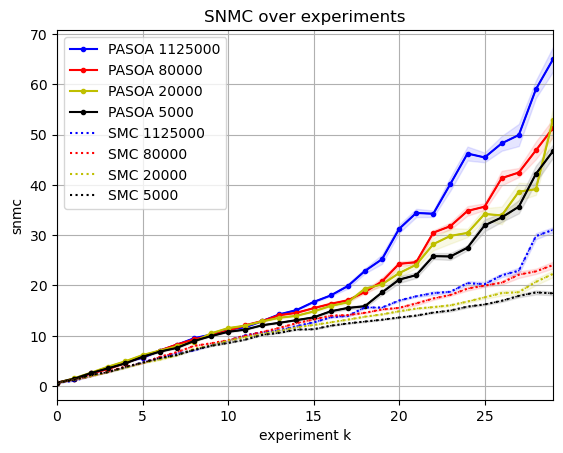}\\
    \includegraphics[trim={.3cm 0 0.2cm 0},clip, height=3cm, width=0.49\linewidth]{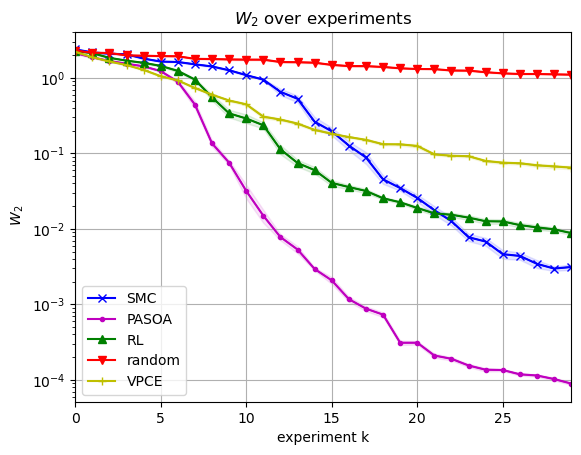}
    \includegraphics[trim={0.7cm 0 0.2cm 0},clip, height=3cm, width=0.49\linewidth]{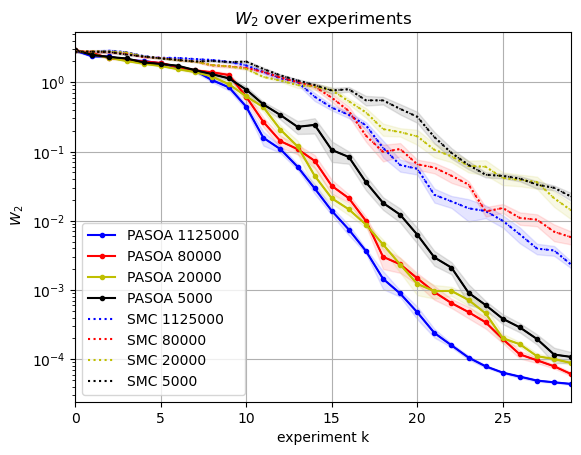}
    \caption{\footnotesize Source location. Column 1: median {and standard error} over 100 rollouts for SPCE (top), SNMC (middle) and L$_2$ Wasserstein distance (bottom) with respect to the number of experiments $k$. Column 2:
    impact of the number of particles (5K to 1M) on  median SPCE, SNMC and Wasserstein distance for PASOA (plain) and SMC (dotted). Note the logarithmic scale.}
    \label{fig:cumul_eig}
\end{figure}

The whole experiment is  repeated  100 times but drawing  source locations at random each time.
Figure \ref{fig:cumul_eig}, column 1,  shows, with respect to $k$, the median {and standard error} for SPCE, SNMC and the L$_2$ Wasserstein distances between weighted particles and the true source locations.

A first observation is that design optimization leads to a significant improvement over the naive random baseline.
 For VPCE and RL-BOED we recover the results shown in Figure 1 of \citet{Blau2022}.
Our method leads to a significant improvement,  both in terms of information gain and posterior estimation. It improves by $30\%$  RL-BOED results on SPCE and  provides much higher SNMC. The L$_2$ Wasserstein distance is two order of magnitude lower, suggesting the higher quality of our measurements.

Both plain SMC and PASOA outperform the other methods.
Additional illustration of the benefit of tempering is visible in
Figure \ref{fig:left4} where PASOA  better estimates the posterior concentrating around the true sources when SMC leads to more scattered particles and is less efficient in removing particles from non informative locations. Tempering allows to reduce the number of particles, outperforming SMC even with much fewer particles, {\it e.g.} $5\mathrm{e}^3$ {\it vs} 10$^6$  for a lower computation time (Figure \ref{fig:cumul_eig}, column 2, and Table \ref{tab:example}). The number of tempering steps tends to decrease with the number of experiments as more information is gathered (Appendix Figure \ref{fig:Tstep}).  PASOA is also more robust to prior misspecification (Figure \ref{fig:miss}).

\begin{table}[h!]
\caption{\footnotesize Impact of the number of particles on average computation times for Source location on a V100 GPU and on average total tempering steps for PASOA. }
\vskip 0.15in
  \label{tab:example}
  \centering
  \resizebox{\linewidth}{!}{%
  \begin{tabular}{|c|c|c|c|c|c|}
    \hline
    Particles  & {50 000}  & 100 000 & 320 000  & 1 000 000 \\
    \hline
        SMC  &  25.1s $\pm$ 0.1s  & 30.1s $\pm$ 0.1s & 61.7 $\pm$ 0.1s  & 130.1 $\pm$ 0.2s \\
    \hline
    PASOA  &  25.9s $\pm$ 0.1s  & 31.4s $\pm$ 0.1s & 63.4 $\pm$ 0.1s  & 134.4 $\pm$ 0.2s \\
    \hline
    Tempering steps  & 120.2 & 120.0 & 116.1 & 114.2    \\
    \hline
  \end{tabular}
  }
\end{table}

%\begin{table}[h!]
%\caption{\footnotesize Impact of the number of particles on average computation times on a V100 GPU, Apple M1Pro CPU and on average total tempering steps for PASOA. }
%\vskip 0.15in
%  \label{tab:example}
%  \centering
%  \resizebox{\linewidth}{!}{%
%  \begin{tabular}{|c|c|c|c|c|c|}
%    \hline
%    Particles  & {5000}  & 20000 & 320000  & 1 000 000 \\
%    \hline
%        SMC (CPU) &  35.5 $\pm$ 0.1s & 142.5 $\pm$ 0.1s & ---  & --- \\
%        SMC (GPU) &  --- & --- & 100.8 $\pm$ 0.1s  & 216.1 $\pm$ 0.1s \\
%    \hline
%    PASOA (CPU) &  35.5 $\pm$ 0.1s & 148.2 $\pm$ 0.1s & ---  & --- \\
%    PASOA (GPU) &  --- & --- &  101.5 $\pm$ 0.1s  & 232.1 $\pm$ 0.1s \\
%    \hline
%    Tempering steps  & 120.2 & 120.0 & 116.1 & 114.2    \\
%    \hline
%  \end{tabular}
%  }
%\end{table}

\begin{figure}
    \centering
    \includegraphics[trim={0.7cm 0 0.2cm 0}, clip,  width=0.49\linewidth]{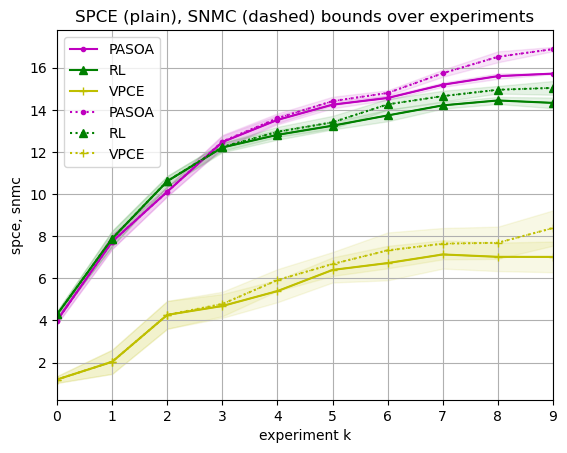}
    \includegraphics[trim={0.7cm 0 0.2cm 0}, clip,  width=0.49\linewidth]{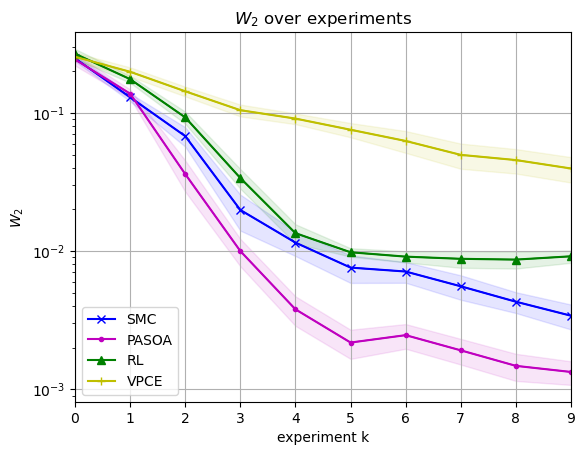}
    \caption{\footnotesize CES example. Median { and standard error} over 100 rollouts, with respect to the number of experiments $k$, for SPCE -plain , SNMC -dashed  (left) and Wasserstein distance (right).
    \label{fig:spce-CES}}
\end{figure}

%\vspace{-0.5cm}
\paragraph{Constant Elasticity of Substitution (CES).}
In this other model \cite{Blau2022,Foster2020}, an agent compares two baskets of goods with 3 items each and gives a rating in $[0,1]$.
The design is a 6-dimensional vector representing quantities for each item in each basket.
 There are 3 parameters  $\thetab=(\rho, \alphab=(\alpha_1,\alpha_2,\alpha_3), u)$ in dimension 5, which have to be recovered from the agent's ratings of different baskets. The design task is challenging since most basket configurations provide very little information. Figure \ref{fig:spce-CES} shows the obtained SPCE, SNMC and Wasserstein distance curves for $K=10$ experiments. SMC-based methods outperform again  RL-BOED and VPCE, both in terms of design sequences and Wasserstein distances. For PASOA, the  [SPCE, SNMC] intervals  (Figure \ref{fig:spce-CES} left)  are distinctly above the other ones. For clarity sake, the SMC [SPCE, SNMC] intervals are  not plotted but shown in Appendix Figure \ref{fig:comp_bounds}.

 For a global view of the  methods, Table \ref{tab:times} summarizes the main features and  running times, with more explanations in Appendix Section \ref{sec:impl}. Additional experimental results  and implementation details are also given in the Appendix.
 The chosen models are benchmarks used in BOED. A real-word application would be a great addition but is out of the scope of this paper.

\begin{table}[h!]
\caption{\label{tab:times} \footnotesize Main features of the compared methods. %\textcolor{blue}{By-product estimation of a posterior distribution (column 2), amortization (column 3) and  non-myopic (column 4) properties.
Column  5 shows training times required for amortization.}
\vskip 0.15in
\centering
  \resizebox{\linewidth}{!}{%
\begin{tabular}{ |c|c|c|c|c| }
\hline
Method & Posterior & Amortized & Non-myopic  & Training Time \\
\hline
PASOA&  \greencheck & \rcross & \rcross
& ---\\
\hline
SMC &  \greencheck & \rcross & \rcross
& ---\\
\hline
RL-BOED \cite{Blau2022} &  \rcross & \greencheck &\greencheck

& \begin{tabular}{c c}CES: $\sim20$h \\Sources: $\sim10$h\end{tabular}\\
\hline
VPCE \cite{Foster2020} &  \greencheck & \rcross  & \rcross

& ---\\
\hline
\end{tabular}
}
\end{table}

\section{Conclusion}
We have introduced a new  Bayesian  sequential design optimization algorithm which also provides  posterior distribution  estimates for parameter inference.
The procedure uses a tempering principle to handle the fact that maximizing information gain jeopardizes standard SMC samplers accuracy.
Although greedy, our approach performs better than the long-sighted reinforcement learning approach of \citet{Blau2022}. As already observed by \citet{Blau2022}, a possible explanation is the use of posterior information in our optimisation process. Moreover, the lesser performance of VPCE, which uses  suboptimal variational posterior approximations, suggests that the key to a good EIG optimization is an accurate posteriors estimation, which can be achieved via SMC and further improved with tempering.
Accurately estimating posterior distributions seems more important and more critical than planning multiple steps ahead.
In addition, when the design parameter is high dimensional, all methods are likely to suffer from the difficulty of searching in a high dimensional design space, but as a myopic approach PASOA is likely to suffer less than non myopic approaches for which the search space grows exponentially as the number of experiments increases.
Nevertheless, investigating the possibility to extend our approach via a policy-based method would be interesting.
Another difficulty may come from a high dimensional parameter. Augmenting the parameter dimension makes inference more difficult. If posterior inference fails, PASOA may loose its specific advantage and result in design of lesser quality. Methods that do not rely on these posterior estimation are less prone to this loss of design performance. However, the goal of experimental design is to provide informative data to infer and ultimately gain information on the parameter of interest. Even if other methods may then provide  better SPCE values, this is useless if the posterior is impossible to estimate correctly.  In practice, for PASOA,  one might use a more sophisticated tuned Markov kernel and exploit the advantage of tempering that a good kernel allows to scale to higher dimensions, see Section 17.2.3 in \cite{chopin2020introduction}.
At last, investigating amortized simulation-based inference such as \cite{Ivanova2021,Kleinegesse2021,Kleinegesse2020} to handle models which are only available through simulations would be an important next step for applications.

\section*{Acknowledgements}
The authors thank the reviewers and area chair for their interesting and useful comments and the Inria Challenge project ROAD-AI for partial funding.
This work was performed using HPC/AI resources from GENCI-IDRIS (Grant 2023-AD011014217R1).
Pierre Alliez is supported by the French government, through the 3IA Côte d’Azur Investments in the Future project managed by the National Research Agency ANR-19-P3IA-0002.

\section*{Impact Statement}
This paper presents work whose goal is to advance the field of Machine Learning. There are many potential societal consequences of our work, none which we feel must be specifically highlighted here.

%\newpage
\bibliographystyle{icml2024}
\bibliography{biblioBOED}

%%%%%%%%%%%%%%%%%%%%%%%%%%%%%%%%%%%%%%%%%%%%%%%%%%%%%%%%%%%%%%
\newpage
\appendix
\onecolumn
%%%%%%%%%%%%%%%%%%%%%%%%%%%%%%%%%%%%%%%%%%%%%%%%%%%%%%%%%%%%%%
\setcounter{proposition}{0}
\setcounter{assumption}{0}

In this appendix, we provide proofs for the theoretical results in our paper and additional details on the implementation and numerical experiments.

\section{Particle EIG contrastive bound optimization}

The algorithm below  summarizes the procedure developed for design optimization. It is the result of the combination of Algorithms \ref{alg:sgd} and \ref{alg:SMC} in the main body. It results from the replacement in the former of the current posterior $p(\thetab | \Db_{k-1})$ by its particle approximation resulting from the adaptive tempering in Algorithm \ref{alg:SMC}.

%\LinesNumbered
\begin{algorithm}
	\caption{Particle EIG contrastive bound stochastic optimization at step $k+1$}
    \label{alg:sgdsmc}
	\begin{algorithmic}[1]
		\State Set $T$ iterations, $\xib_0$, stepsizes $(\gamma_t)_{t=1:T}$ %\\
        \State Run tempered SMC Algorithm \ref{alg:SMC} (main body) at step $k$
    for $M=N(L+1)$,
   %with multinomial resampling,
   and set $L+1$ independent particle approximations $P_{k,\ell}^N$, for $\ell=0:L$  (see main paper Section \ref{Sec:combi} for details)
		 \While{$t \leq T$}
		  \State Sample  $\thetab^i_{\ell, t} \! \sim P_{k,\ell}^N,  \mbox{for $\ell\!= \!0\! :\! L, i\!=\!1\!: \!N_t$ }$
		\State Sample $u^i_{t} \sim  p(u),  \quad \mbox{ for $i=1: N_t$}$
      \State Set  $\!\nabla_{\!t+1} \!\!= \!\!\frac{1}{N_t} \!\!\sum\limits_{i=1}^{N_t} \!\!\nabla_\xib \!F(\xib,\!  \thetab^i_{0t},  \cdot ,\! \thetab^i_{Lt}, \!T^{\xib}_{\thetab^i_{0t}}\!\!(u^i_{t})\!)|_{\xib = \xib_t}\!\!$
        \State Update $ \xib_{t+1} = \xib_t + \gamma_t  {\nabla}_{t+1}$
    \EndWhile
	  \State \Return{$\xib_{k+1,N}^*\!=\!\xib_T$ or a Polyak averaging value}
  \end{algorithmic}
\end{algorithm}

\section{Differentiation under the integral sign}
\label{sec:deriv}

We briefly recall standard conditions under which it is possible to exchange differentiation and expectation operators. In our setting this implies conditions that are also useful for other results in Section \ref{sec:cons}. Indeed to carry out a stochastic gradient algorithm (Algorithms \ref{alg:sgd} and \ref{alg:sgdsmc}), we assume that all quantities are well defined. In particular,
\begin{align}
 \nabla_{\!\xib}\!I_{\!PCE}(\xib) \!=  \! \Exp_{p(u)\prod_{\ell=0}^L p(\thetab_\ell)}\!\left[ \nabla_{\!\xib} F(\xib, \! \thetab_0,  \cdot, \!\thetab_L,\! T^{\xib}_{\thetab_0}\!(U))\right].  \label{def:gradipceapp}
 \end{align}
 Denoting $p_u \otimes p_\theta^{\otimes L+1}$ the product probability distribution $p(u)\prod_{\ell=0}^L p(\thetab_\ell)$ on ${\cal U} \times \Thetab^{L+1}$, sufficient conditions  for (\ref{def:gradipceapp}) are that the function $F$ is differentiable in $\xib$ for $p_u \otimes p_\theta^{\otimes L+1}$-almost all $(u, \thetab_0, \cdot , \thetab_L)$ and its gradient satisfies $\left|\nabla_{\!\xib} F(\xib, \! \thetab_0,  \cdot, \!\thetab_L,\! T^{\xib}_{\thetab_0}\!(u))\right| \leq H(\thetab_0, \cdot , \thetab_L,u)$ where $H$ has a bounded expectation. In particular this implies the continuity properties $(ii)$ and $(iii)$ required in Lemma \ref{lem:hypA1} and \ref{lem:hypA2} below.

In practice, it can also be useful to notice that the gradient required within the expectation (\ref{def:gradipceapp}) can be expressed using only the  log-likelihood gradient. To simplify, in the expression below we denote $T^{\xib}_{\thetab_0}(u)$ by $y$. Using the definition of $F$, it comes,
\begin{align}
  \nabla_{\!\xib} F(\xib, \! \thetab_0,  \cdot, \!\thetab_L, y ) &=  \nabla_{\!\xib} \log p( y | \xib, \thetab_0)  - \sum_{\ell=0}^{L} w_\ell(\thetab_0, \cdot, \thetab_L,\xib,y) \; \nabla_{\!\xib} \log p( y | \xib, \thetab_\ell)\label{def:gradsimplif}
 \end{align}
 where the $w_\ell(\thetab_0, \cdot, \thetab_L,\xib,y) = \frac{p( y | \xib, \thetab_\ell)}{\sum\limits_{\ell'=0}^L p( y | \xib, \thetab_{\ell'})}$ sum to 1 and act as weights.
 Thus we only need to compute for all $\thetab_0,\thetab_\ell$, the gradient
 $ \nabla_{\!\xib} \log p( T^{\xib}_{\thetab_0}(u) | \xib, \thetab_\ell)$.

\section{Proofs of main and intermediate results}

We first specify the steps that lead to  Proposition \ref{propL2} in the main body. The overall goal is to show that products of particle approximations have similar convergence properties  than single particle approximations.

\subsection{Product form estimators}
\label{sec:prodapp}

Product form estimators have been used in various places in the literature but only studied in a more formal and general way in some recent work \cite{Kuntz2022}. Their advantages are clearly highlighted in \cite{Kuntz2022} but the theoretical results therein do not cover the use of product form estimators within SMC samplers. In contrast, \cite{Rebeschini2015,Lindsten2017,Kuntz2023,Aitchison2019}  consider SMC settings that may look similar to ours but differ in key aspects.
Indeed, \cite{Rebeschini2015} proposed a so-called block particle filter consisting of a product of local smaller particle filters to approximate particle filters in high dimensional models. The last line of their Algorithm 2 clearly shows that the block particle filter is equivalent to running several independent particle filters on smaller dimensional spaces. In contrast,  we run a single SMC  from which we extract at each step a product form estimate of the target distribution. This allows to compute expectations and other inferential quantities in a more efficient way, using the advantages of product form estimators, notably their lower variance, as explained in \cite{Kuntz2022}.
The Divide-and-Conquer SMC (DaC-SMC) methodology introduced in \cite{Lindsten2017}
and further studied in \cite{Kuntz2023} targets a different objective, which is to generalize the sequential nature of standard SMC to a more general tree-structured execution flow of the sampler. Hierarchical decompositions of variables of interest into subsets are considered and product form estimators are used to recombined the splitted subsets of particles. Typical use cases include sets of variables exhibiting a graph dependence structure such as Markov random fields.
In contrast, in our work, the execution flow remains sequential, corresponding to a tree that is a simple line, and we do not address complex parameter structures.
 At last, the Tensor Monte Carlo technique proposed in \cite{Aitchison2019} shares similarity with DaC-SMC but focuses on computational aspects and does not cover as many theoretical properties. The work in \cite{Aitchison2019}
 proposes to exploit dependence structures in variables and conditional independence properties to address the issue of computing the very large sums induced by product form estimators. This is not an issue we encounter as, in our work,  the computation of very large sums is avoided via stochastic approximation.

 The advantage of using a product of particle approximations justifies a new analysis of the SMC procedure.
Interestingly, the product-form can also help to mitigate the growth of constant $c_k$
in the L$_2$ bound in Proposition \ref{propL2} ($c_\tau'$ in the proof below). Increasing $L$ can allow to reduce $c_k$ to an extent that can make the bound informative for finite sample $N$. See the remark at the end of Section \ref{sec:conv}.

\subsection{Convergence of product form particle approximations}
\label{sec:conv}

 The result is intuitive and generalizes standard SMC proofs but as mentioned in the previous section, we believe it is not covered in previous work. It requires additional care as illustrated in Lemma \ref{lem:varsum} below. We follow the presentation in Chapter 11 of \citet{chopin2020introduction} which is mainly based on the presentation in \cite{Crisan2002}. As mentioned in \cite{chopin2020introduction} it is not the most general presentation but it has the advantage of using standard tools while presenting the general idea. For simplicity, we only present the L$_2$ convergence result, which is enough to prove Proposition \ref{propL2}.

To simplify the notation, we use $\tau$ as the time or step index and  $\mu_\tau$, $M_\tau$ for $, \mu_{\lambda_\tau}, M_{\lambda_\tau}$ as in Algorithm \ref{alg:SMC}, to denote the step  and successive targets  and Markov kernels, but this extends to $k$ and $p_k$, as we repeat the same process sequentially.
For a SMC procedure on $\Thetab$,  $G_\tau$ denotes then the so-called potential function involved in the weights normalization and
$\tilde{G}_\tau = G_\tau /(\mu_{\tau-1} M_{\tau-1}(G_\tau))$ its normalization. For example, in our setting in  Section 6, $G_\tau(\thetab) = G_{k,\tau}(\thetab) = p(\yv_k | \thetab, \xib_k)^\gamma$. Recall that  for a function $\phi$ on $\Thetab$,  we then have
\begin{align}
\mu_\tau[\phi] = \Exp_{\mu_\tau}[\phi] &=\mu_{\tau-1}M_{\tau-1}[\phi \; \tilde{G}_\tau]= \int \tilde{G}_\tau(\thetab) \phi(\thetab) \left\{ \int M_{\tau-1}(\thetab',\thetab)\;  \mu_{\tau-1}(\thetab')\;  d\thetab' \right\} d\thetab . \label{def:rec1}
\end{align}
As a particular case, we have
\begin{align}
&\mu_\tau(\thetab_\ell) = \mu_\tau[\delta_{\thetab_\ell}] = \tilde{G}_\tau(\thetab_\ell) \left\{ \int M_{\tau-1}(\thetab',\thetab_\ell) \; \mu_{\tau-1}(\thetab')\;  d\thetab' \right\} \; d\thetab  \label{def:rec1dirac} \\
&\mu_{\tau-1}M_{\tau-1}[\tilde{G}_\tau] =  \mu_\tau[1] = 1
\label{def:un}
\end{align}

We generalize these definitions and results on the product space $\Thetab^{L+1}$ with $\phi$ being now a function of $L+1$ variables.
For a function $\phi$ on $\Thetab^{L+1}$, we use notation $M_\tau\phi$ for the function
\begin{align}
    &M_\tau\phi: (\thetab_0, \cdot, \thetab_L) \rightarrow \int \cdot  \int  \phi(\thetab'_0,\cdot,\thetab'_L) \left(\prod_{\ell=0}^L M_\tau(\thetab_\ell, \thetab'_\ell)\;  d\thetab'_\ell \right). \label{def:Mphi}
    \end{align}
Similarly, we denote $\mu_\tau^{\otimes L+1}[\phi] = \Exp_{\mu_\tau^{\otimes L+1}}\!\left[\phi\right]= \int \cdot  \int  \phi(\thetab_0,\cdot,\thetab_L) \left(\prod_{\ell=0}^L \mu_\tau(\thetab_\ell) \; d\thetab_\ell \right) .$
Defining,
\begin{align*}
(\mu_{\tau-1}M_{\tau-1})^{\otimes L+1}[\phi] &= \int \cdot  \int  \phi(\thetab_0, \cdot,  \thetab_L)
 \prod\limits_{\ell=0}^L \left\{\int M_{\tau-1}(\thetab'_\ell, \thetab_\ell)\; \mu_{\tau-1}(\thetab'_\ell) \; d\thetab'_\ell  \right\} d\thetab_\ell .
 %\label{def:recL2}
\end{align*}
we can deduce from (\ref{def:rec1}) (respectively via (\ref{def:rec1dirac}) and (\ref{def:un}) that
\begin{align}
&\mu_\tau^{\otimes L+1}[\phi ] = (\mu_{\tau-1}M_{\tau-1})^{\otimes L+1}\!\left[\phi \prod_{\ell=0}^L \tilde{G}_\tau(\thetab_\ell)\right],  \label{def:recL1} \\
&(\mu_{\tau-1}M_{\tau-1})^{\otimes L+1}\!\left[\prod_{\ell=0}^L \tilde{G}_\tau(\thetab_\ell)\right] = \mu_\tau^{\otimes L+1}[1]=1 . \label{def:recL12}
\end{align}

The strategy is to establish Proposition \ref{propL2} by decomposing
 the error at step $\tau$ as a sum of contributions from each step (sampling,
reweighting, resampling, etc.) of the current and previous  steps.
We first need a general result on variances of sum of random variables.
\begin{lem} \label{lem:varsum}
Let $\{X_\ell^i\}_{\ell=0:L, i=1:N}$ a collection of {\it i.i.d.} random variables on ${\cal X}$. For any function $\phi$ on ${\cal X}^{L+1}$ so that the variances exist we have, for all $(i_0, \cdot , i_L) \in [1:N]^{L+1}$,
$$Var\left(\phi(X_0^{i_0}, \cdot, X_L^{i_L}) \right) = Var\left(\phi(X_0^{1}, \cdot , X_L^{1}) \right) $$
and we denote this common variance by $Var(\phi_1)$.
We can then bound the following variance
\begin{align*}
Var\left(\sum_{i_0=1}^N \cdot  \sum_{i_L=1}^N \phi(X_0^{i_0}, \cdot, X_L^{i_L}) \right) \leq N^{L+1} \left( N^{L+1}-(N-1)^{L+1}\right)  Var(\phi_1)
\end{align*}
\end{lem}
\paragraph{Proof.}
Let us denote $I=[1:N]$, $J = I^{L+1}$, $j=(i_0, \cdot ,i_L)$ for $i_\ell \in I$, and use the short notation $\phi_j = \phi(X_0^{i_0}, \cdot , X_L^{i_L})$. Then
\begin{align*} Var\left(\sum_{i_0=1}^N \cdot \sum_{i_L=1}^N \phi(X_0^{i_0}, \cdot , X_L^{i_L}) \right)& =  Var(\sum\limits_{j \in J} \phi_j) = \sum_{j\in J} Var(\phi_j) + \sum\limits_{j\in J} \sum_{j'\not= j} Cov(\phi_{j'}, \phi_j) \\
&= N^{L+1} Var(\phi_1) +  \sum\limits_{j\in J} \sum_{j'\not= j} Cov(\phi_{j'}, \phi_j) \; .
\end{align*}
The second covariance term is a sum of $N^{2(L+1)} - N^{L+1}$ pairwise covariances among which a number are zero due to the independence of the $\{X_\ell^i\}_{\ell=0:L, i=1:N}$. To count them, we notice
that for each $j=(i_0, \cdot , i_L) \in J$, $Cov(\phi_j, \phi_j') =0$ for every $j' \in J$ such that $j'= (i'_0, \cdot , i'_L)$ with $i'_\ell \in I \backslash \{i_\ell\}$. There are thus $(N-1)^{L+1}$ such $j'$. It follows that there are altogether $N^{L+1} (N-1)^{L+1}$ null covariance terms and  $N^{L+1}(N^{L+1}-(N-1)^{L+1}-1)$ non null ones. For each non null covariance, the Cauchy-Schwartz inequality implies that $ Cov(\phi_{j'}, \phi_j) \leq \sqrt{Var(\phi_{j'}) Var(\phi_j)} = Var(\phi_1)$, which concludes the proof.
\hfill
$\square$

\vspace{.3cm}

We then recall some notation and definitions specific to our product form particle approximation.
The tempered SMC procedure in Algorithm \ref{alg:SMC} produces $M=N(L+1)$ particles denoted by $\thetab_{\tau}^{1:M}$ at each step indexed by $\tau$. When partitioning these particles into $L+1$ disjoint subset we will denote  by $\thetab_{\ell,\tau}^{1:N}$ the $N$ particles in each subset with $\ell=0:L+1$.
We denote by $\zeta_{k,L}^{N}$ the collection of random variables  $\zeta_{k,L}^{N} = \{W_{k,\ell}^{i},\thetab^{i}_{k,\ell}\}_{i=1:N, \ell=0:L}$, and use the short notation $p_k^{\otimes L+1}[\phi]$ for $p_k^{\otimes L+1}[\phi] = \Exp_{\prod_{\ell=0}^L p(\thetab_\ell|\Db_k)}[\phi(\thetab_0, \cdot, \thetab_L)]$.
We write ${\cal F}_{\tau-1}$ for  the $\sigma$-algebra ${\cal F}_{\tau-1} = \sigma\!\left(\thetab^{1:M}_{\tau-1}\right)$.
Let ${\cal C}_b(\Thetab^{L+1})$
denote the set of functions $\phi: \Thetab^{L+1} \rightarrow \Rset$  that are measurable and bounded, and
let  $||\phi ||_\infty$ denote the supremum norm $||\phi ||_\infty = \sup_{\thetab \in \Thetab^{L+1}} |\phi(\thetab)|$.
The following lemma bounds the Monte Carlo error at step $\tau$. %{\it i.e.} the error related to sampling $M$ particles from this conditional distribution.

\begin{lem}[\bf Monte Carlo error] \label{lem:MCE}
Using Algorithm \ref{alg:SMC} with a multinomial resampling scheme, for all functions $\phi \in {\cal C}_b(\Thetab^{L+1})$,
\begin{align} \Exp_{\zeta_{k,L}^{N}}\!\left[\left\{ \frac{1}{N^{L+1}} \sum_{i_0=1}^N \cdot \sum_{i_L=1}^N  \phi(\thetab^{i_0}_{0,\tau},\cdot, \thetab^{i_L}_{L,\tau}) -  \!\!
\sum_{i_0=1}^N\cdot  \sum_{i_L=1}^N \!\left(\prod_{\ell=0}^L W_{\ell, \tau-1}^{i_\ell}\right)\!  M_{\tau-1}\phi(\thetab^{i_0}_{0, \tau-1}, \cdot, \thetab^{i_L}_{L, \tau-1}) \right\}^2 \right] \!\! \leq  \! c_{N,L} \Vert\phi \Vert_\infty^2
 \label{eq:lemMCE}
\end{align}
with $c_{N,L}= 1- \left(1-\frac{1}{N}\right)^{L+1}$ and
where the expectation is taken over all the realizations of the random tempered SMC method, or equivalently on $\zeta_{k,L}^{N}$.
\end{lem}

\paragraph{Proof.}
Multinomial resampling preserves conditional independence so that the particles  $\thetab_{\ell,\tau}^{1:N}$ are {\it i.i.d.} conditionally on ${\cal F}_{\tau-1}$ and their conditional distribution is
$$\thetab^i_{\ell,\tau} | {\cal F}_{\tau-1} \sim  \sum_{i=1}^N W^i_{\ell,\tau-1} M_{\tau-1}(\thetab_{\ell, \tau-1}^i, \cdot).$$
Thus for all $i_{0}, \cdot , i_L$,
\begin{align*}
\Exp_{\zeta_{k,L}^{N}}\!\left[ \phi(\thetab^{i_0}_{0,\tau},\cdot, \thetab^{i_L}_{L,\tau})  \left\vert \right. {\cal F}_{\tau-1}\right] &= \sum_{i_0=1}^N \cdot  \sum_{i_L=1}^N    \left(\prod_{\ell=0}^L W_{\ell, \tau-1}^{i_\ell}\right)  M_{\tau-1}\phi(\thetab^{i_0}_{0,\tau-1}, \cdot, \thetab^{i_L}_{L,\tau-1}) \\
&=  \Exp_{\zeta_{k,L}^{N}}\!\left[ \frac{1}{N^{L+1}} \sum_{i_0=1}^N \cdot  \sum_{i_L=1}^N  \phi(\thetab^{i_0}_{0,\tau},\cdot, \thetab^{i_L}_{L,\tau})  \left\vert\right.  {\cal F}_{\tau-1} \right] .
\end{align*}
It follows that
\begin{align*}
&\Exp_{\zeta_{\tau,L}^{N}}\!\left[\!\left\{ \frac{1}{N^{L+1}} \!\!\sum_{i_0=1}^N \cdot  \sum_{i_L=1}^N  \phi(\thetab^{i_0}_{0,\tau},\cdot, \thetab^{i_L}_{L,\tau})\! -  \!
\sum_{i_0=1}^N \cdot  \sum_{i_L=1}^N   \!\!  \left(\prod_{\ell=0}^L W_{\ell, \tau-1}^{i_\ell}\right)  M_{\tau-1}\phi(\thetab^{i_0}_{0, \tau-1}, \cdot, \thetab^{i_L}_{L, \tau-1}) \right\}^2  \!   \left\vert\right. {\cal F}_{\tau-1} \!\right]  \\
& = Var\!\left[ \frac{1}{N^{L+1}} \sum_{i_0=1}^N \cdot  \sum_{i_L=1}^N  \phi(\thetab^{i_0}_{0,\tau},\cdot , \thetab^{i_L}_{L,\tau})   \left\vert\right. {\cal F}_{\tau-1} \right]
 \; =\;  \frac{1}{N^{2(L+1)}} Var\!\left[\sum_{i_0=1}^N \cdot  \sum_{i_L=1}^N  \phi(\thetab^{i_0}_{0,\tau},\cdot , \thetab^{i_L}_{L,\tau})   \left\vert\right. {\cal F}_{\tau-1} \right] \\
&\leq  c_{N,L} \; Var\!\left[  \phi(\thetab^{1}_{0,\tau},\cdot , \thetab^{1}_{L,\tau})    \left\vert\right.  {\cal F}_{\tau-1} \right] \\
&  \leq   c_{N,L} \;  \Exp_{\zeta_{\tau,L}^{N}}\!\left[  \phi(\thetab^{1}_{0,\tau},\cdot , \thetab^{1}_{L,\tau})^2    \left\vert\right. {\cal F}_{\tau-1} \right] \leq c_{N,L} \; \Vert \phi\Vert^2_\infty
\end{align*}
The line before last above results from Lemma \ref{lem:varsum}.
Using the tower property, we get (\ref{eq:lemMCE}).
\hfill
$\square$

\vspace{.3cm}

The constant $c_{N,L}$ is equivalent to $\frac{L+1}{N}$ and tends to 0 when $N$ tends to $\infty$.
The next Lemma involves the potential functions $G_{\tau}$.
Note that in Section 6, $G_{\tau}$ corresponds to $G_{k,\tau}(\thetab)= p(\yv_k | \thetab, \xib)^{\gamma}$, where $\gamma$ usually also depends on $\tau$ and $k$ as it is found adaptively.
%not numbered anymore: by solving equation (11) in the paper.
However all we need is that the potential and then the likelihood is upper bounded in $\thetab$.

\begin{lem}[\bf Weights normalization error] \label{lem:G}
If $G_{\tau}$ is upper bounded, then for all functions $\phi \in {\cal C}_b(\Thetab^{L+1})$,
\begin{align*}
\Exp_{\zeta_{\tau,L}^{N}}\!\left[\left\{\sum_{i_0=1}^N \cdot \sum_{i_L=1}^N  \left(\prod\limits_{\ell=0}^L W_{\ell,\tau}^{i_\ell} \right) \phi(\thetab^{i_0}_{0,\tau},\cdot, \thetab^{i_L}_{L,\tau}) -  \frac{1}{N^{L+1}}
\sum_{i_0=1}^N\cdot  \sum_{i_L=1}^N \!\left(\prod_{\ell=0}^L \tilde{G}_{\tau}(\thetab_{\ell,\tau}^{i_\ell})\right)\! \phi(\thetab^{i_0}_{0,\tau}, \cdot, \thetab^{i_L}_{L,\tau}) \right\}^2 \right]  \\
 \leq   \Vert\phi \Vert_\infty^2 \; \Exp_{\zeta_{k,L}^{N}}\!\left[\left\{  \prod\limits_{\ell=0}^L \left( \frac{1}{N} \sum_{i=1}^N \tilde{G}_{\tau}(\thetab_{\ell,\tau}^i)\right) -1 \right\}^2 \right]
\end{align*}
where $\tilde{G}_{\tau} = G_{\tau}/(\mu_{\tau-1} M_{\tau-1}(G_\tau))$.
\end{lem}

\paragraph{Proof.}

By definition
$$ W_{\ell,\tau}^{i_\ell}  = \frac{{G}_{\tau}(\thetab_{\ell,\tau}^{i_\ell})}{\sum_{i=1}^N {G}_{\tau}(\thetab_{\ell,\tau}^i)}
= \frac{\tilde{G}_{\tau}(\thetab_{\ell,\tau}^{i_\ell})}{\sum_{i=1}^N \tilde{G}_{\tau}(\thetab_{\ell,\tau}^i)}$$
Thus the term in the square in the left hand side is
$$\left(\sum_{i_0=1}^N \cdot \sum_{i_L=1}^N  \left(\prod\limits_{\ell=0}^L W_{\ell,\tau}^{i_\ell} \right) \phi(\thetab^{i_0}_{0,\tau},\cdot, \thetab^{i_L}_{L,\tau})\right)  \left(1- \frac{1}{N^{L+1}}\prod\limits_{\ell=0}^L (\sum_{i=1}^N \tilde{G}_{\tau}(\thetab_{\ell,\tau}^i))  \right) $$ and the first factor is bounded by $\Vert \phi \Vert_{\infty}$ since $\sum\limits_{i_0=1}^N \cdot \sum\limits_{i_L=1}^N  \left(\prod\limits_{\ell=0}^L W_{\ell,\tau}^{i_\ell} \right) =1$.
\hfill
$\square$

\vspace{.3cm}

The two Lemmas can now be combined to establish the proposition below.
\begin{proposition}[\bf L$_2$ convergence] \label{prop:L22}
Using Algorithm \ref{alg:SMC} with multinomial resampling and assuming that potential functions ${G}_{\tau}$ are upper bounded, there exists constants $c_\tau,c'_\tau$ so that,  for all functions $\phi \in {\cal C}_b(\Thetab^{L+1})$,
\begin{align}
 \Exp_{\zeta_{k,L}^{N}}\!\left[\left\{ \frac{1}{N^{L+1}} \sum_{i_0=1}^N \cdot \sum_{i_L=1}^N  \phi(\thetab^{i_0}_{0,\tau},\cdot, \thetab^{i_L}_{L,\tau}) -  \!\! (\mu_{\tau-1}M_{\tau-1})^{\otimes L+1}[\phi] \right\}^2 \right] \!\! \leq  \! c_\tau \; c_{N,L} \; \Vert\phi \Vert_\infty^2
 \label{eq:prop21} \\
 \Exp_{\zeta_{k,L}^{N}}\!\left[\left\{ \sum_{i_0=1}^N \cdot \sum_{i_L=1}^N  \left(\prod\limits_{\ell=0}^L W_{\ell,\tau}^{i_\ell} \right)   \phi(\thetab^{i_0}_{0,\tau},\cdot, \thetab^{i_L}_{L,\tau}) -  \!\! \mu_{\tau}^{\otimes L+1}[\phi] \right\}^2 \right] \!\! \leq  \! c'_\tau\;  c_{N,L} \; \Vert\phi \Vert_\infty^2
 \label{eq:prop22}
\end{align}
with $c_{N,L} = 1 - (1-\frac{1}{N})^{L+1}$.
\end{proposition}

\paragraph{Proof}
The proof works by induction on $\tau$.
Replacing $\mu_{\tau-1} M_{\tau-1}$ by $\mu_0$, at $\tau=0$, (\ref{eq:prop21}) holds  with $c_0=1$.
Assume (\ref{eq:prop21}) holds at step $\tau$, we first show that   (\ref{eq:prop22}) also holds at step $\tau$. The left hand side in (\ref{eq:prop22}) can be decomposed into
\begin{align*}
&\sum_{i_0=1}^N \cdot \sum_{i_L=1}^N  \left(\prod\limits_{\ell=0}^L W_{\ell,\tau}^{i_\ell} \right)   \phi(\thetab^{i_0}_{0,\tau},\cdot, \thetab^{i_L}_{L,\tau}) -   \mu_{\tau}^{\otimes L+1}[\phi] =  \\
&\sum_{i_0=1}^N \cdot \sum_{i_L=1}^N  \left(\prod\limits_{\ell=0}^L W_{\ell,\tau}^{i_\ell} \right)   \phi(\thetab^{i_0}_{0,\tau},\cdot, \thetab^{i_L}_{L,\tau}) -    \frac{1}{N^{L+1}}
\sum_{i_0=1}^N\cdot  \sum_{i_L=1}^N \!\left(\prod_{\ell=0}^L \tilde{G}_{\tau}(\thetab_{\ell,\tau}^{i_\ell})\right)\! \phi(\thetab^{i_0}_{0,\tau}, \cdot, \thetab^{i_L}_{L,\tau})  \\
+ &  \frac{1}{N^{L+1}}
\sum_{i_0=1}^N\cdot  \sum_{i_L=1}^N \!\left(\prod_{\ell=0}^L \tilde{G}_{\tau}(\thetab_{\ell,\tau}^{i_\ell})\right)\! \phi(\thetab^{i_0}_{0,\tau}, \cdot, \thetab^{i_L}_{L,\tau})    -   \mu_{\tau}^{\otimes L+1}[\phi] \; .
\end{align*}
For  all $\phi$, when $\phi'(\thetab_0, \cdot, \thetab_L) =  \left(\prod_{\ell=0}^L \tilde{G}_{\tau}(\thetab_{\ell})\right)\! \phi(\thetab_0, \cdot, \thetab_L)$, we have
$(\mu_{\tau-1}M_{\tau-1})^{\otimes L+1}[\phi']= \mu_{\tau}^{\otimes L+1}[\phi]$ by (\ref{def:recL1}). The second term can then be bounded by $c_\tau c_{N,L} \Vert\phi' \Vert_\infty^2$,    applying (\ref{eq:prop21}) with $\phi'$.
For the first term, we apply Lemma \ref{lem:G} and then (\ref{eq:prop21}) with $\phi(\thetab_0, \cdot, \thetab_L)= \prod_{\ell=0}^L \tilde{G}_{\tau}(\thetab_{\ell})$ since then $ (\mu_{\tau-1}M_{\tau-1})^{\otimes L+1}[\phi]= \mu_\tau^{\otimes L+1}[1]=1 $ by (\ref{def:recL12}).
Finally we get (\ref{eq:prop22}) with $c'_\tau=  4 c_\tau \Vert \tilde{G}_\tau\Vert_\infty^{2(L+1)}$ where we have used that $\Exp[(X+Y)^2] \leq 2 (\Exp[X^2]+ \Exp[Y^2])$.
We then show that  (\ref{eq:prop22})  at step $\tau-1$ implies  (\ref{eq:prop21})  at step $\tau$.
Similarly, we have,
\begin{align*}
&\frac{1}{N^{L+1}} \sum_{i_0=1}^N \cdot \sum_{i_L=1}^N  \phi(\thetab^{i_0}_{0,\tau},\cdot, \thetab^{i_L}_{L,\tau}) -   (\mu_{\tau-1}M_{\tau-1})^{\otimes L+1}[\phi] =  \\
&\frac{1}{N^{L+1}} \sum_{i_0=1}^N \cdot \sum_{i_L=1}^N  \phi(\thetab^{i_0}_{0,\tau},\cdot, \thetab^{i_L}_{L,\tau}) -  \sum_{i_0=1}^N \cdot \sum_{i_L=1}^N  \left(\prod\limits_{\ell=0}^L W_{\ell,\tau-1}^{i_\ell} \right)   M_{\tau-1}\phi(\thetab^{i_0}_{0,\tau-1},\cdot, \thetab^{i_L}_{L,\tau-1})  \\
+ &   \sum_{i_0=1}^N \cdot \sum_{i_L=1}^N  \left(\prod\limits_{\ell=0}^L W_{\ell,\tau-1}^{i_\ell} \right)   M_{\tau-1}\phi(\thetab^{i_0}_{0,\tau-1},\cdot, \thetab^{i_L}_{L,\tau-1})  -  (\mu_{\tau-1}M_{\tau-1})^{\otimes L+1}[\phi] \; .
\end{align*}

The first term can be bounded using Lemma \ref{lem:MCE} and for the second term by applying (\ref{eq:prop22}) at $\tau-1$ to function $M_{\tau-1}\phi$ defined in (\ref{def:Mphi}). Then (\ref{eq:prop21}) holds with $c_\tau= 2(1 +c'_{\tau-1} ) $.
\hfill
$\square$

\vspace{.3cm}

Proposition \ref{propL2} in the paper is just the result  above but reduced to the presentation of (\ref{eq:prop22}).
To give more insight on the constants involved in Proposition \ref{propL2}, we note that the constant denoted by $c_k$ corresponds to $c_\tau'$ in the above proof. A well referenced behavior in SMC, see {\it e.g.} \cite{chopin2020introduction}, is that this constant may increase very fast with  $k$ and become uninformative for finite $N$.
Without considering tempering for simplicity, it comes that $c_0=1$ and $c_k= 4(2+c_{k-1}) \Vert \tilde{G}\Vert_\infty^{2(L+1)}$, where $\Vert \tilde{G}\Vert_\infty$ denotes the upper bound in $\thetab$ of $ \tilde{G}_k(\thetab) \propto p(\yv_k  | \thetab, \xib_k)$, which is assumed here to be independent on the specific value of $\yv_k$  and $\xib_k$.
It follows that $c_k= 2 \sum_{i=1}^k (4\Vert \tilde{G}\Vert_\infty^{2(L+1)})^i$.
Interestingly, with our product form approximation, if $\Vert \tilde{G}\Vert_\infty <1$,  $c_k$ remains low and decreases when $L$ increases.

%In our work, we limited the presentation to {\it simple} conditions but with stronger assumptions, in particular on the Markov kernels, it is possible to limit the growing of $c_k$ over steps $k$, see Section 11.4. in \cite{chopin2020introduction}.}

\subsection{Consistency of the sequential design estimators}
\label{sec:cons}

We specify the steps that lead to the proof of Proposition \ref{prop:consTSMC} in the main body. We first recall some notation and definitions.
The sequential design  values  produced by Algorithm \ref{alg:sgdsmc} above can be seen as realizations of random estimators $\{\xib_{k+1,N}^*\}_{N\geq 1}$ targeting points of maximum of random criterion functions $\{I_{PCE}^{k+1,N}(\xib)\}_{N\geq 1}$,
\begin{align*}
I^{k+1,N}_{\!PCE}(\xib) &=  \Exp_{p(u)\prod_{\ell=0}^L {p}^N_{k,\ell}(\thetab_\ell)}\!\left[ F(\xib, \! \thetab_0,  \cdot, \!\thetab_L,\! T^{\xib}_{\thetab_0}\!(U))\right] \; .
\end{align*}
A natural question is to study the limiting distributions of these random quantities when the number of particles $N$ tends to infinity. The main property is the following Theorem \ref{prop:unif}, which results from  a general result from M-estimator theory \cite{vaart_1998,Vaartwellner1996}. The two following lemmas are then useful to provide simpler sufficient conditions to satisfy the Theorem's assumptions and to establish our main result in Proposition \ref{prop:consTSMC} in the main body, also recalled below.

\begin{theorem}[\bf Theorem 5.7 in \cite{vaart_1998}] \label{prop:unif}
Assume
\begin{itemize}
\item[]{(i)} For all $\epsilon>0$,
\begin{align}
 \lim_{N\rightarrow \infty} p_ {\zeta_{k,L}^{N}}\!\!\!\left(\sup_{\xib \in {\cal E}} |{I}^{k+1,N}_{\!PCE}(\xib) - I^{k+1}_{\!PCE}(\xib)| \geq \epsilon\right) =0. \label{Ai}
\end{align}
\item[]{(ii)} For all $\epsilon>0$,
\begin{align}
\sup_{\Vert \xib-\xib^*_{k+1}\Vert\geq \epsilon} I^{k+1}_{\!PCE}(\xib) < I^{k+1}_{\!PCE}(\xib^*_{k+1}) \; . \label{Aii}
\end{align}
\item[]{(iii)} There exists a sequence of positive random variables $\{\rho_N\}_{N\geq 1}$ and a sequence of random variables  $\{ \xib^*_{k+1,N}\}_{N\geq 1}$ in ${\cal E}$ that satisfy
$$\forall \epsilon>0, \quad \lim_{N\rightarrow \infty} p_ {\zeta_{k,L}^{N}}\!\!\!\left(\rho_N \geq \epsilon \right) =0$$
$$  \lim\inf_{N\rightarrow \infty} p_ {\zeta_{k,L}^{N}}\!\!\!\left( {I}^{k+1,N}_{\!PCE}(\xib^*_{k+1,N}) \geq  I^{k+1,N}_{\!PCE}(\xib^*_{k+1}) -\rho_N  \right) =1.$$
\end{itemize}
Then the sequence of estimators  $\{ \xib^*_{k+1,N}\}_{N\geq 1}$ is consistent, {\it i.e.} for all $\epsilon >0$,
$$  \lim_{N\rightarrow \infty} p_ {\zeta_{k,L}^{N}}\!\!\!\left(\Vert \xib^*_{k+1,N} - \xib^*_{k+1}\Vert \geq \epsilon\right) =0. $$
\end{theorem}

\paragraph{Proof.}
The proof is a special case of Theorem 5.7 in \cite{vaart_1998}. We reproduce it using our notation.
In all the following proofs, to simplify, we drop the $k+1$ notation, so that $I^{k+1}_{\!PCE}(\xib)$ and $I^{k+1,N}_{\!PCE}(\xib)$ are now simply denoted by $I_{\!PCE}(\xib)$ and $I^{N}_{\!PCE}(\xib)$ and their respective maximizers by $\{ \xib^*\}$ and $\{ \xib^*_{N}\}$.
Since $\xib^*$  maximizes $I_{\!PCE}(\xib)$, it comes that for all $N \geq 1$,
\begin{align*}
0 & \leq I_{\!PCE}(\xib^*) - I_{\!PCE}(\xib^*_N)  \\
& = (I_{\!PCE}(\xib^*) -  I^N_{\!PCE}(\xib^*)) +(I^N_{\!PCE}(\xib^*) - I^N_{\!PCE}(\xib^*_N)) +  (I^N_{\!PCE}(\xib^*_N)-  I_{\!PCE}(\xib^*_N)) \; .
\end{align*}
The first and third terms in the sum are bounded by $\sup\limits_{\xib \in {\cal E}} | I_{\!PCE}(\xib) -  I^N_{\!PCE}(\xib)|$ while the second term can be bounded
by $\rho_N + (I^N_{\!PCE}(\xib^*) - I^N_{\!PCE}(\xib^*_N) - \rho_N) \; \delta_{\{I^N_{\!PCE}(\xib^*) - I^N_{\!PCE}(\xib^*_N) > \rho_N\}}$ where $\delta_{{\cal A}}$ is the indicator function which is 1 if ${\cal A}$ is satisfied and 0 otherwise.
It follows that
\begin{align*}
0 &\leq I_{\!PCE}(\xib^*) - I_{\!PCE}(\xib^*_N)   \\
& \leq 3 \max\!\left(\!2 \sup\limits_{\xib \in {\cal E}} | I_{\!PCE}(\xib) -  I^N_{\!PCE}(\xib)|, \rho_N,  \left(I^N_{\!PCE}(\xib^*) - I^N_{\!PCE}(\xib^*_N) - \rho_N\right)\!\delta_{\{I^N_{\!PCE}(\xib^*) - I^N_{\!PCE}(\xib^*_N) > \rho_N\}}\!\!\right).
\end{align*}
Assumptions $(i)$  and $(iii)$ imply that all three terms in the max tend to 0 in probability so that for all $\eta >0$,
\begin{align}
  \lim_{N\rightarrow \infty} p_ {\zeta_{k,L}^{N}}\!\!\!\left(I_{\!PCE}(\xib^*) - I_{\!PCE}(\xib^*_N)   \geq \eta \right) =0. \label{theocvp}
  \end{align}
Then, for all $\epsilon > 0$, using $(ii)$,  for all $\xib$ satisfying $\Vert \xib-\xib^*\Vert\geq \epsilon$,  there exists $\eta >0$ so that $I_{\!PCE}(\xib) \leq I_{\!PCE}(\xib^*)  - \eta$. This implies that
$\{\Vert \xib^*_N-\xib^*\Vert\geq \epsilon\} \subset \{ I_{\!PCE}(\xib^*_N) \leq I_{\!PCE}(\xib^*)  - \eta \}$ and
$$  p_ {\zeta_{k,L}^{N}}\!\!\!\left( \Vert \xib^*_N-\xib^*\Vert\geq \epsilon \right) \leq p_ {\zeta_{k,L}^{N}}\!\!\!\left(I_{\!PCE}(\xib^*_N) \leq I_{\!PCE}(\xib^*)  - \eta \right) = p_ {\zeta_{k,L}^{N}}\!\!\!\left(   I_{\!PCE}(\xib^*) - I_{\!PCE}(\xib^*_N) \geq \eta \right).$$
The limit in $N$ of this last term tends to 0 by (\ref{theocvp}),  which concludes the proof.
\hfill
$\square$

\vspace{.3cm}

In general, the uniform convergence in $(i)$ is the most difficult assumption to check but in our setting, when ${\cal E}$ is compact, it is easily derived from previous assumptions and the pointwise convergence in probability ($(iv)$ below), which can be derived from the L$_2$ convergence in Proposition \ref{propL2} (see comments in the main body). Recall the following definition,
$$f_{PCE}(\xib,\thetab_0, \cdot, \thetab_L)=   \Exp_{p(u)}\left[F(\xib, \! \thetab_0,  \cdot, \!\thetab_L,\! T^{\xib}_{\thetab_0}\!(U))\right] $$ and the shortcut notation
$p_k^{\otimes L+1}= \prod\limits_{\ell=0}^L p(\thetab_\ell|\Db_k)$.
\begin{lem} \label{lem:hypA1}
Assume
\begin{itemize}
\item[]{(i)} ${\cal E} \in \Rset^d$ is a compact set.
\item[]{(ii)} $I^{k+1}_{\!PCE}(\xib)$ is a continuous function in $\xib$.
\item[]{(iii)} $f_{PCE}(\xib,\thetab_0, \cdot , \thetab_L)$ is a continuous function in $\xib$ for $p_{k}^{\otimes L+1}$-almost all $(\thetab_0, \cdot , \thetab_L)$.
\item[]{(iv)} Convergence in probability pointwise: For all $\xib \in {\cal E}$ and all $\epsilon>0$,
$$ \lim_{N\rightarrow \infty} p_ {\zeta_{k,L}^{N}}\!\!\!\left( |{I}^{k+1,N}_{\!PCE}(\xib) - I^{k+1}_{\!PCE}(\xib)| \geq \epsilon\right) =0. $$
\end{itemize}
Then the uniform convergence in Theorem \ref{prop:unif}~$(i)$  is satisfied, that is, for all $\epsilon>0$,
\begin{align*}
 \lim_{N\rightarrow \infty} p_ {\zeta_{k,L}^{N}}\!\!\!\left(\sup_{\xib \in {\cal E}} |{I}^{k+1,N}_{\!PCE}(\xib) - I^{k+1}_{\!PCE}(\xib)| \geq \epsilon\right) =0.
\end{align*}
\end{lem}

\paragraph{Proof.}
Continuous functions on a compact set are uniformly continuous. It follows from $(ii)$ and $(iii)$ that for all $\epsilon >0$, there exists $\eta>0$ so that for all $\xib' \in {\cal E}$,
\begin{equation}
\sup_{\Vert \xib - \xib'\Vert \leq \eta} |I_{PCE}(\xib) - I_{PCE}(\xib')| \leq \epsilon \label{eq:unicon1}
\end{equation}
and for $p_{k}^{\otimes L+1}$-almost all $(\thetab_0, \cdot , \thetab_L)$,
\begin{equation}
\sup_{\Vert \xib - \xib'\Vert \leq \eta} |f_{PCE}(\xib,\thetab_0, \cdot , \thetab_L) - f_{PCE}(\xib',\thetab_0, \cdot , \thetab_L)| \leq \epsilon . \label{eq:unicon2}
\end{equation}
Let ${\cal B}(\xib,\eta)$ be a ball centered at $\xib$ with radius $\eta$.  As ${\cal E}$ is compact, for all $\eta>0$, it is possible to extract, from the cover set $\bigcup\limits_{\xib \in{\cal E}} {\cal B}(\xib, \eta)$,  a finite subcover $\bigcup\limits_{b=1:B} {\cal B}(\xib^{(b)}, \eta)$ so that ${\cal E} \subset \bigcup\limits_{b=1:B} {\cal B}(\xib^{(b)}, \eta)$ and
\begin{eqnarray}
\sup_{\xib \in {\cal E}} |{I}^{N}_{\!PCE}(\xib) - I_{\!PCE}(\xib)| &=& \max\limits_{b=1:B} \sup\limits_{\xib \in {\cal B}(\xib^{(b)}, \eta)} |{I}^{N}_{\!PCE}(\xib) - I_{\!PCE}(\xib)| \;.\label{eq:maxsup}
\end{eqnarray}
For all $b=1:B$, and all $\xib \in {\cal B}(\xib^{(b)}, \eta)$,
we also have that
$$ |{I}^{N}_{\!PCE}(\xib) - I_{\!PCE}(\xib)| \leq  |{I}^{N}_{\!PCE}(\xib) - I^N_{\!PCE}(\xib^{(b)})|+ |{I}^{N}_{\!PCE}(\xib^{(b)}) - I_{\!PCE}(\xib^{(b)})|+ |{I}_{\!PCE}(\xib^{(b)}) - I_{\!PCE}(\xib)|$$
which implies
\begin{eqnarray}
 \sup\limits_{\xib \in {\cal B}(\xib^{(b)}, \eta)} |{I}^{N}_{\!PCE}(\xib) - I_{\!PCE}(\xib)| &\leq & \sup\limits_{\xib \in {\cal B}(\xib^{(b)}, \eta)} \left\{ |{I}^{N}_{\!PCE}(\xib) - I^N_{\!PCE}(\xib^{(b)})| \right\} \label{eq:majo} \\
 &+ &  |{I}^{N}_{\!PCE}(\xib^{(b)}) - I_{\!PCE}(\xib^{(b)})| \nonumber\\
 &+&
\sup\limits_{\xib \in {\cal B}(\xib^{(b)}, \eta)}  \left\{ |{I}_{\!PCE}(\xib^{(b)}) - I_{\!PCE}(\xib)| \right\} \nonumber
\end{eqnarray}

For the first term in the right-hand side,
$$\sup\limits_{\xib \in {\cal B}(\xib^{(b)}, \eta)}\!\! \left\{ |{I}^{N}_{\!PCE}(\xib) - I^N_{\!PCE}(\xib^{(b)})| \right\} \leq \Exp_{p_{k}^{N\otimes L+1}}\!\!\left[ \sup\limits_{\Vert \xib - \xib^{(b)}\Vert \leq \eta} |f_{PCE}(\xib,\thetab_0, \cdot , \thetab_L) - f_{PCE}(\xib^{(b)},\thetab_0, \cdot , \thetab_L)|  \right]$$
By Proposition \ref{prop:L22}~(\ref{eq:prop22}), the right-hand side above tends in L$_2$-norm and then in probability when $N$ tends to $\infty$ to
$$\Exp_{p_{k}^{\otimes L+1}}\!\!\left[ \sup\limits_{\Vert \xib - \xib^{(b)}\Vert \leq \eta} |f_{PCE}(\xib,\thetab_0, \cdot , \thetab_L) - f_{PCE}(\xib^{(b)},\thetab_0, \cdot , \thetab_L)|  \right],$$ which is smaller than $\epsilon$ by (\ref{eq:unicon2}).
The second term in the right-hand side of (\ref{eq:majo}) tends in probability to 0 by $(iv)$ for all $\xib^{(b)}$. Finally,
using (\ref{eq:unicon1}), the third term  in (\ref{eq:majo}) is smaller than $\epsilon$.
To conclude, (\ref{eq:maxsup}) implies the uniform convergence (\ref{Ai}).
\hfill
$\square$

\vspace{.3cm}

The following result gives simpler sufficient conditions for Assumption $(ii)$ in Theorem \ref{prop:unif} to hold.
\begin{lem} \label{lem:hypA2}
Assume
\begin{itemize}
\item[]{(i)} ${\cal E} \in \Rset^d$ is a compact set.
\item[]{(ii)} $I^{k+1}_{\!PCE}(\xib)$ is a continuous function in $\xib$ .
\item[]{(iii)} For all $\xib \not = \xib^*_{k+1}$, $I^{k+1}_{\!PCE}(\xib) < I^{k+1}_{\!PCE}(\xib^*_{k+1})$ .
\end{itemize}
Then Assumption  $(ii)$ of Theorem \ref{prop:unif} is satisfied.
\end{lem}

\paragraph{Proof.}
A continuous function reaches its maximum on a compact set. It follows that $I^{k+1}_{\!PCE}(\xib)$ reaches its maximum on  the compact subset ${\cal E}\backslash {\cal B}(\xib^*_{k+1},\epsilon)$. Let $\xib_\epsilon$ denote a value at which this maximum is reached. Then $(iii)$ implies that $I^{k+1}_{\!PCE}(\xib_\epsilon) < I^{k+1}_{\!PCE}(\xib^*_{k+1})$, which leads to (\ref{Aii}).
\hfill
$\square$

\vspace{.3cm}

Using Lemmas \ref{lem:hypA1} and \ref{lem:hypA2}, we can now replace in Theorem \ref{prop:unif} Assumptions $(i)$ and $(ii)$ by simpler conditions. It follows the Proposition \ref{prop:consTSMC} presented in the paper and recalled below.
\begin{proposition} \label{prop:consTSMCapp}
Assume
\begin{assumption} ${\cal E} \in \Rset^d$ is a compact set.
\end{assumption}
\begin{assumption} For all $\xib \not = \xib^*_{k+1}$, $I^{k+1}_{\!PCE}(\xib) < I^{k+1}_{\!PCE}(\xib^*_{k+1})$
\end{assumption}
\begin{assumption} There exists a sequence of positive random variables $\{\rho_N\}_{N\geq 1}$ and a sequence of random variables  $\{ \xib^*_{k+1,N}\}_{N\geq 1}$ in ${\cal E}$ that satisfy
$$\forall \epsilon>0, \quad \lim\limits_{N\rightarrow \infty} p_ {\zeta_{k,L}^{N}}\!\!\!\left(\rho_N \geq \epsilon \right) =0$$
$$  \lim\inf\limits_{N\rightarrow \infty} p_ {\zeta_{k,L}^{N}}\!\!\!\left( {I}^{k+1,N}_{\!PCE}(\xib^*_{k+1,N}) \geq  I^{k+1,N}_{\!PCE}(\xib^*_{k+1}) -\rho_N  \right) =1.$$
\end{assumption}
Then the sequence of estimators  $\{ \xib^*_{k+1,N}\}_{N\geq 1}$ is consistent, {\it i.e.} for all $\epsilon >0$,
$$  \lim\limits_{N\rightarrow \infty} p_ {\zeta_{k,L}^{N}}\!\!\!\left( \Vert \xib^*_{k+1,N} - \xib^*_{k+1}\Vert \geq \epsilon\right) =0. $$
\end{proposition}

 \paragraph{Proof.}
 With ${\cal E}$ compact, we can use Lemma \ref{lem:hypA1}. The continuity of $I^{k+1}_{\!PCE}(\xib)$ and $f_{PCE}$ has been already assumed earlier as specified in Section \ref{sec:deriv} and Lemma \ref{lem:hypA1}~$(iv)$ is a consequence of Proposition \ref{prop:L22}. It follows the uniform convergence property $(i)$ in Theorem \ref{prop:unif}. Then
(A1-2) and Lemma \ref{lem:hypA2} imply $(ii)$ in Theorem \ref{prop:unif}.
 With (A3), Theorem \ref{prop:unif} leads to the result.
Note that if we assume that $\xib^*_{k+1,N}$ is an exact maximizer of ${I}^{k+1,N}_{\!PCE}(\xib) $ then (A3) is trivially satisfied with $\rho_N=0$.

\section{Numerical experiments}
\label{sec:expapp}

\subsection{Sequential prior contrastive estimation (SPCE) criterion}
\label{sec:spce}

We specify the SPCE  introduced by \citet{Foster2021} and used in our experiments and those of \citet{Blau2022} to assess the design sequence quality in our comparison.
For a number $K$ of experiments, $\Db_K= \{(\yv_1,\xib_1), \cdot, (\yv_K,\xib_K) \}$ and $L$ contrastive variables, SPCE is defined
as
\begin{align}
    SPCE(\xib_1, \cdot, \xib_K) &= \Exp_{\prod\limits_{k=1}^K p(\yv_k | \xib_k, \thetab_0) \; \prod\limits_{\ell=0}^L p(\thetab_\ell)}\left[\log \frac{\prod\limits_{k=1}^K p(\yv_k | \thetab_0, \xib_k)}{\frac{1}{L+1}\sum\limits_{\ell=0}^L \prod\limits_{k=1}^K p(\yv_k | \thetab_\ell, \xib_k)} \right] \; . \label{def:spce}
\end{align}
SPCE is a lower bound of the total EIG which is the expected
information gained from the entire sequence of design parameters $\xib_1, \ldots, \xib_K$ and it becomes  tight when $L$ tends to $\infty$.  In addition,
SPCE has the advantage to use only samples from the prior $p(\thetab)$ and not from the successive posterior distributions. It makes it a fair criterion to compare methods on design  sequences only.
Considering a true parameter value denoted by $\thetab^*$, given a sequence of design values $\{\xib_k\}_{k=1:K}$, observations $\{\yv_k\}_{k=1:K}$ are simulated using $p(\yv | \thetab^*, \xib_k)$ respectively. Therefore, for a given $\Db_k$, the corresponding SPCE is estimated numerically by sampling $\thetab_1, \cdot, \thetab_L$ from the prior,
\begin{align*}
    SPCE(\Db_K) &= \frac{1}{N} \sum\limits_{i=1}^N  \left\{\log \frac{\prod\limits_{k=1}^K p(\yv_k | \thetab^*, \xib_k)}{\frac{1}{L+1}\left(\prod\limits_{k=1}^K p(\yv_k | \thetab^*, \xib_k)+ \sum\limits_{\ell=1}^L \prod\limits_{k=1}^K p(\yv_k | \thetab^{i}_\ell, \xib_k)\right)} \right\} \; .
\end{align*}

As shown in \cite{Foster2021} (Appendix A), SPCE increases with $L$ to reach the total EIG $I(\xib_1, \ldots, \xib_K)$ when $L\rightarrow \infty$ at a rate ${\cal O}(L^{-1})$ of convergence. More specifically, it is shown in \cite{Foster2021} that
\begin{align}
0\leq  I(\xib_1, \ldots, \xib_K) - SPCE(\xib_1, \ldots, \xib_K) \leq \frac{C}{L+1} \label{enc:spce}
\end{align}
where $C = \Exp_{p(\Db_K) p(\thetab | \Db_K)}\left[\frac{p(\Db_K  | \thetab)}{p(\Db_K)} \right] -1 $ with the notation $p(\Db_K  | \thetab) = \prod\limits_{k=1}^K p(\yv_k | \thetab, \xib_k)$.

It is also shown in \cite{Foster2021} that for a given $L$, SPCE is bounded by $\log(L+1)$ while the upper bound SNMC below is potentially unbounded. As in \cite{Blau2022}, if  we use $L=10^7$ to compute SPCE and SNMC, the bound is $\log(L+1) = 16.12$ for SPCE. In practice this does not impact the numerical methods comparison as the intervals [SPCE, SNMC] containing the total EIG remain clearly distinct.

\subsection{Sequential nested Monte Carlo (SNMC) criterion}
\label{sec:snmc}

Similarly, an upper bound on the total EIG,  with similar features, has also been introduced by \citet{Foster2021}. Its expression is very similar to that of SPCE, varying only through the sum in the denominator,

\begin{align*}
    SNMC(\xib_1, \cdot, \xib_K) &= \Exp_{\prod\limits_{k=1}^K p(\yv_k | \xib_k, \thetab_0) \; \prod\limits_{\ell=0}^L p(\thetab_\ell)}\left[\log \frac{\prod\limits_{k=1}^K p(\yv_k | \thetab_0, \xib_k)}{\frac{1}{L}\sum\limits_{\ell=1}^L \prod\limits_{k=1}^K p(\yv_k | \thetab_\ell, \xib_k)} \right] \; .
\end{align*}

 \subsection{Implementation details}
 \label{sec:impl}

For VPCE \cite{Foster2020} and RL-BOED \cite{Blau2022}, we use the code available at \href{https://github.com/csiro-mlai/RL-BOED/tree/master}{github.com/csiro-mlai/RL-BOED}, using the settings recommended therein to reproduce the results in the respective papers.
From the obtained sequences of observations and design values, we compute SPCE and SNMC as explained above and retrieve the same results as in their respective papers.

Our code is implemented in Jax \cite{jax2018github} and available at \href{https://github.com/iolloj/pasoa}{github.com/iolloj/pasoa}. Several packages are used through the repository. Namely, we used Optax \cite{deepmind2020jax} to run Gradient Descents, the Sequential Monte Carlo part was heavily inspired and built using Kernels from BlackJax \cite{blackjax2020github} and we used OTT \cite{cuturi2022optimal} to compute Wasserstein distances.

Table \ref{tab:timesapp} summarizes the main features and running times of the compared methods. The RL-BOED method has the advantage to be both non-myopic and amortized in the sense that a policy is learnt upfront and then used straightforwardly at each new experiment. It follows a much longer training time, which does not exist for the other methods. Note that in comparison the deployment times of all methods are neglible (see Table \ref{tab:example} in the paper). In contrast RL-BOED does not provide approximations for the posterior distributions.

\begin{table}[h!]
\caption{\label{tab:timesapp} Main features and training times of the compared methods: the second column indicates whether a method also provides approximation of posterior distribution, the third if it is amortized and the fourth if it is non-myopic. The last column shows training times for the amortized method RL-BOED and a sequence of $K$ experiments run on a single Nvidia V100 GPU,  for the source finding and CES examples.}
\vskip 0.15in
\centering
\begin{tabular}{ |c|c|c|c|c| }
\hline
Method & Posterior & Amortized & Non-myopic  & Training Time \\
\hline
PASOA&  \greencheck & \rcross & \rcross
& ---\\
\cline{1-5}
SMC &  \greencheck & \rcross & \rcross
& ---\\
\cline{1-5}
RL-BOED \cite{Blau2022} &  \rcross & \greencheck &\greencheck
& \begin{tabular}{c c}CES: $\sim20$h \\Sources: $\sim10$h\end{tabular}\\
\cline{1-5}
VPCE \cite{Foster2020} &  \greencheck & \rcross  & \rcross &
 ---\\
\hline

\hline
\end{tabular}
\end{table}

 \subsection{Hardware details}

Our method can be run on a local machine and was tested on a Apple M1 Pro 16Gb chip. However, for a faster running time, each experiment was finally produced by running our method on a single Nvidia V100 GPU. One other advantage of tempering and of our PASOA method is that by reducing the number of needed particles for an accurate procedure, it lowers the hardware requirements for this method as it becomes feasible to run it on CPUs (see Table \ref{tab:example}).

\subsection{Checking the assumptions given in the theoretical results}

Ideally, the models used in experiments should satisfy the assumptions appearing in our propositions.
For the L$_2$ convergence result (Proposition \ref{propL2}), the conditions are easy to check. Proposition \ref{propL2} requires that the potential functions $G_{k,\tau}$ are bounded. It is sufficient to check that the likelihood  $p(\yv | \thetab, \xib)$ as a function of $\thetab$ is bounded (main Section 6 before Proposition \ref{propL2}). For the source location model, the likelihood is log-normal and is bounded independently of $\thetab,\yv$, and  $\xib$. For the CES example, the likelihood is a mixture given in equation (\ref{eq:mix}), Section \ref{sec:ces} below where the last component is a logit-normal distribution. Both $p_0$ and $p_1$ in equations (\ref{eq:p0}) and (\ref{eq:p1}) below are in $[0,1]$. The only potentially problematic case may be when $\sigma_\eta \rightarrow 0$. In that case, $1-p_0$ tends to 0 and for $p_1$ we can use the approximation below equation (\ref{eq:p1}). It follows that the third term
$(1-p_0-p_1)\; q(\yv | \thetab, \xib)$ in (\ref{eq:mix})
remains bounded.

For the consistency result (Proposition \ref{prop:consTSMC}), conditions (A1) and (A2) can be stronger than necessary. Note that condition (A3) is not directly related to the model but to the optimization procedure and could be ignored. The important {\it weaker} condition is (ii) in Theorem \ref{prop:unif}. Similarly to consistency results in M-estimator theory (see {\it e.g.} \citet{vaart_1998}), in our work, we assume that
$\xib_{k+1}^*$
is a global and unique maximum of $I_{PCE}^{k+1}$ ((A2)). Condition (ii) actually states that $\xib_{k+1}^*$
is in addition well separated (see Figure  5.2 of \citet{vaart_1998} for an illustration of this notion). Lemma \ref{lem:hypA2} gives sufficient conditions for (ii), which results in (A1) and (A2) in Proposition \ref{prop:consTSMC}. (A1) is that the design space is compact and (A2) states that
$\xib_{k+1}^*$
is a unique global maximum (not necessary well separated). (A1) is easy to check but (A2) is strong and not usually easy to check. Both can be relaxed with additional technicalities, see section 5.2.1 of \citet{vaart_1998}.
For the CES model, the design space is compact. For the source example, it can be restricted to $[-X,X]^2$
without specific care, as {\it e.g}. in \cite{Blau2022}. For (A2), we have not found yet a general way to check this for the $I_{PCE}$ bound. Note though, that this {\it unchecked} assumption is common practice as it would require more care to talk about consistency if the maximum was not unique and global.

\subsection{Source location example}
\label{sec:exapp}

For the 2D  location finding example  used in \cite{Foster2021,Blau2022} and tested in the paper, with 2 sources, $K=30$ successive design optimisations,  and 100 repetitions of the experiment, the number of gradient steps was set to 5000 and the ESS for the SMC procedure to 0.9.
Also, in practice stratified resampling was preferred to multinomial resampling. The latter has the advantage to considerably simplify the proofs and this mismatch between theory and practice is very common in the SMC literature.
 Figure \ref{fig:cumul_eig} in the paper   shows the SPCE, SNMC and the L$_2$ Wasserstein distances  between weighted particles and the true source locations, providing three quantitative assessment and comparison of methods. As an additional, visual assessment of the quality of the posterior approximation provided by our method,
Figure \ref{fig:left4} in the paper and Figure \ref{fig:partic_apprx}  below illustrate the evolution of the particles over the design steps, starting from a sample following the prior to a sample concentrating around the true source locations.
In particular, the $k=0$ step shows particles simulated according to the prior.
In most use cases, plain SMC already gives better results than other reference methods.
 Figure \ref{fig:left4} and Table \ref{tab:example} in the paper show that tempering allows to reduce the number of particles.
In Figure  \ref{fig:partic_apprx} below, the source locations, indicated by red crosses in the plots, are chosen in a part of the space not well covered by the prior to illustrate the robustness of our approach to a potential prior misspecification. After some iterations PASOA is able to explore the parameter space to finally concentrate the posterior on the true source locations.
In contrast, SMC may miss some of the sources when they are outside the prior mass or when there are too many of them.
We suspect that prior misspecification is a typical very common feature that jeopardizes SMC performance while impacting much less PASOA. This is actually the same problem encountered with IS that tempering aims at solving.
Similarly, Figure \ref{fig:miss} shows, in terms of SPCE, SNMC and Wasserstein distance, that SMC is more robust with tempering.
Figure \ref{fig:Tstep}  indicates the number of  tempering steps taken on average.  The median (over 100 rollouts) number of tempering steps varies from 14 to 2 and is globally decreasing, being under 5 after 15 experiments. The number of tempering steps reduces when the posterior concentrates and when adding new observations becomes less informative.

\begin{figure}[ht!]
    \centering

    \includegraphics[width=1\textwidth]{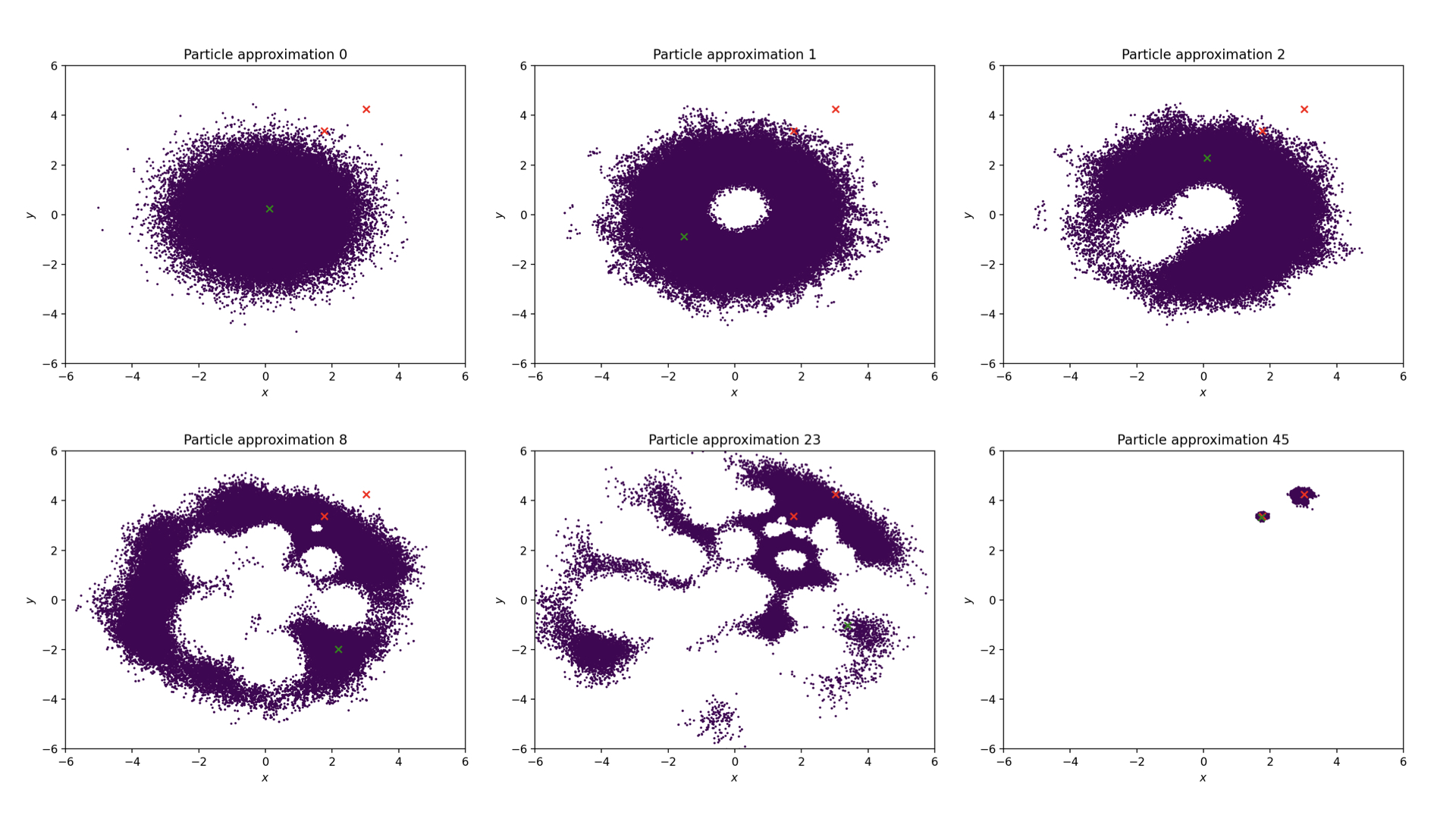}
    \caption{PASOA evolution of particles (in purple) over some selected steps $k$. Particles correspond initially to a sample from the prior $p(\thetab)$ and progressively evolve to a sample of particles located around the initially unknown true source positions indicated by red crosses. Green crosses indicate the optimal measurement locations $\xib_k^*$ obtained at each step $k$. }
    \label{fig:partic_apprx}
\end{figure}

\begin{figure}[h!]
\centering
    \includegraphics[width=0.39\linewidth]{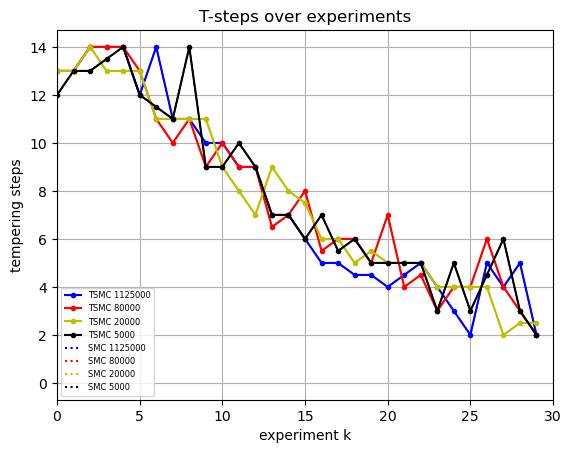}
  \caption{Source location example: median (over 100 rollouts) number of tempering steps over the number of experiments, with respect to the number of particles. }
  \label{fig:Tstep}
\end{figure}

\begin{figure}[h!]
\resizebox{\textwidth}{!}{%
\begin{tabular}{ccc}
    \includegraphics[width=0.33\linewidth]{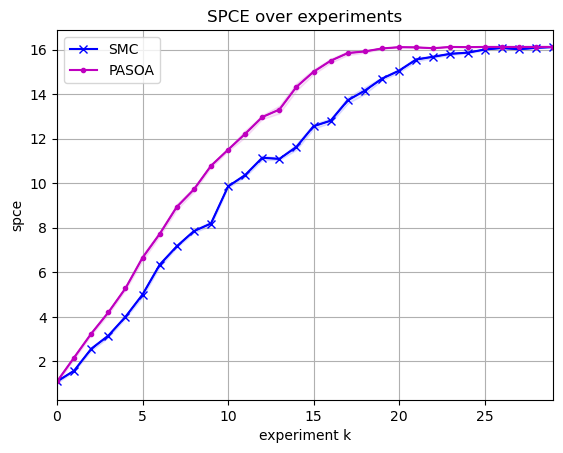} &
  \includegraphics[width=0.33\linewidth]{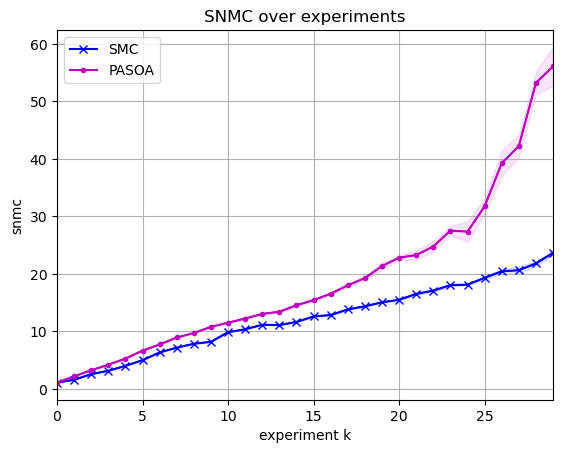} &
    \includegraphics[width=0.33\linewidth]{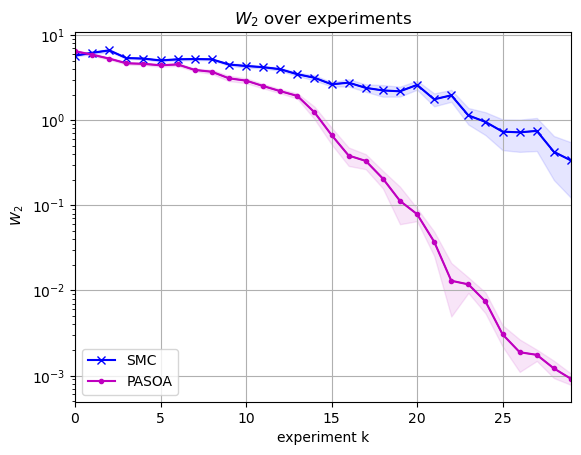}
  \end{tabular}
  }
  \caption{Source location example. Prior misspecification: median (over 100 rollouts) (a) SPCE, (b) SNMC  and (c) Wasserstein distances for SMC (blue) and PASOA (red). }
  \label{fig:miss}
\end{figure}

\subsection{Constant Elasticity of Substitution example}
\label{sec:ces}

This other model, used in \cite{Blau2022,Foster2020}, comes from behavioral economics.
In this model,  an agent compares two baskets of goods and gives a rating $y$ on a sliding scale from 0 to 1.  The goal is to design the two baskets of goods so as to infer the agent's utility function, which depends on some unknown parameters. The designs are 6-dimensional vectors $\xib = (\xib_1, \xib_2)$ corresponding to the two baskets with 3 values each $\xib_d=(\xib_{d,1}, \xib_{d,2}, \xib_{d,3}) \in [0:100]^3$ for $d=1,2$, which represent quantities for 3 items in each basket.
 There are 3 parameters $\thetab=(\rho, \alphab, u)$ in dimension 5, whose prior distributions are respectively $\rho \sim Beta(1,1)$, $\alphab=(\alpha_1,\alpha_2,\alpha_3) \sim Dirichlet(1,1,1)$ and $\log u \sim {\cal N}(1,3)$.

 The model likelihood is given by the following  model that uses a subjective utility function $U$ and two hyperparameters $\epsilon=2^{-22}$ and $ \tau = 0.005$,
 \begin{align*}
  y &= f(\eta, \epsilon) \\
  \mbox{where } & \eta \sim {\cal N}(\mu_\eta, \sigma_\eta^2)  \\
  \mbox{with } & \mu_\eta = (U(\xib_1)- U(\xib_2))\; u \\
     \sigma_\eta &= (1 + \Vert \xib_1 - \xib_2\Vert)\; \tau \; u \\
     \mbox{For $d=1,2$,} \quad U(\xib_d) &= \left(\alpha_1 \xib_{d,1}^{\rho} + \alpha_2 \xib_{d,2}^{\rho} + \alpha_3\xib_{d,3}^{\rho} \right)^{1/\rho}
 \end{align*}
 where $f(\eta, \epsilon)$ takes it values in $[\epsilon, 1-\epsilon]$ and is a censored sigmoid defined by
 \begin{eqnarray*}
     f(\eta, \epsilon) &= &1-\epsilon \quad  \mbox{if $\eta \geq logit(1-\epsilon)$}\\
     &=& \epsilon \quad \mbox{if $\eta \leq logit(\epsilon)$ } \\
      &=& (1+\exp(-\eta))^{-1}  \quad \mbox{otherwise}
 \end{eqnarray*}
 with $logit(y)= \log(y/(1-y))$.
In other words, $y$ is a censored logit-normal distribution with parameters $\mu_\eta$ and $\sigma_\eta$. Its density is a mixture
\begin{align}
p(y | \thetab, \xib) &= p_0 \delta_\epsilon (y)+ p_1 \delta_{1-\epsilon}(y)  + (1-p_0-p_1) \; q(y | \thetab, \xib) \label{eq:mix}
\end{align}
where $q(y | \thetab, \xib)= \frac{1}{\sigma_\eta \sqrt{2\pi} y (1-y)} \exp(\frac{(logit(y)-\mu_\eta)^2}{2\sigma_\eta^2})$ is the density of a logit-normal distribution and $p_0$ and $p_1$ are defined by the following logit-normal CDF values
\begin{eqnarray}
p_0&= &q(y \leq \epsilon) = p(\eta \leq logit(\epsilon)) = F\left(\frac{logit(\epsilon) - \mu_\eta}{\sigma_\eta}\right) \label{eq:p0}\\
p_1 &=&1-q(y \leq 1-\epsilon) = 1- p(\eta \leq logit(1-\epsilon)) = 1 -F\left(\frac{logit(1-\epsilon) - \mu_\eta}{\sigma_\eta}\right) \label{eq:p1}
\end{eqnarray}
 with the last equalities involving the normal CDF values of variable $\eta$ and the standard normal CDF $F$ values.
In practice, computing $\log p_0$ and $\log p_1$ may sometimes be numerically problematic when $p_0$ or $p_1$ become too small. Computing $p_0$ or $p_1$ involves computing lower and upper Gaussian tails. In this case, following \cite{Foster2020}, we use
the following first order asymptotic approximation of the standard normal CDF, when $x$ is large,
$$ 1-F(x) \approx \frac{1}{ x \; \sqrt{2\pi}}  \; \exp(-x^2/2).$$
and when $x$ is small (negative)
$$ F(x) = 1 - F(-x) \approx \frac{1}{-x \sqrt{2\pi}}  \exp(-x^2/2).$$
Thus, denoting $f$ the pdf of the standard normal distribution, $\log p_0 \approx \log f(x) - log(-x)$ with $x=\frac{logit(\epsilon) - \mu_\eta}{\sigma_\eta} $
and
 $\log p_1 \approx \log f(x) - log(x)$ with $x=\frac{logit(1-\epsilon) - \mu_\eta}{\sigma_\eta} $
 or to summarize both approximations when $|x|$ is  large, $\log f(x) - \log(|x|)$.

Implementation details, if not otherwise specified, are the same as for the source location  example.
 We plan $K=10$ successive design optimisations and repeat the whole experiment 100 times for varying values of the true parameters, for all methods, PASOA, SMC, RL-BOED, VPCE and the random design baseline.
This is overall a more challenging example as the objective function has many suboptimal local maxima, and the  stochastic gradient procedure may be more sensitive to initialization.
The number of gradient steps was set to 5000.
For the SMC procedure, the ESS was set to 0.9,   the Markov
kernel is that of a random walk Metropolis-Hasting scheme with
prior transformations mapping the parameters to $\Rset^4$. The transformations used are respectively, $u'=\log u$, $\rho'=logit(\rho)=\log \frac{\rho}{1-\rho}$, and $\alpha_1'=\log \frac{\alpha_1}{\alpha_3}, \alpha_2'=\log \frac{\alpha_2}{\alpha_3}$, with the inverse transformations being $u = \exp u'$, $\rho=\frac{\exp\rho'}{1+\exp \rho'}$, $\alpha_1=\frac{\exp \alpha_1'}{1+\exp \alpha_1'+\exp \alpha_2'}$, $\alpha_2=\frac{\exp \alpha_2'}{1+\exp \alpha_1'+\exp \alpha_2'}$ and $\alpha_3=\frac{1}{1+\exp \alpha_1'+\exp \alpha_2'}$.
We use then $L = 100$ contrastive variables with each $N=500$ simulations.

Figure \ref{fig:spce-CES} in the paper and Figures \ref{fig:comp_bounds} and \ref{fig:ces_all_w} below show, with respect to $k$, the median and standard error of the SPCE, SNMC and  Wasserstein distances between weighted particles and the true parameters.
We observe for all methods more variability in the repetitions for this example. {In terms of total EIG, the difference with RL-BOED is not as large as in the source location example, but the difference remains large for Wasserstein distances. PASOA still shows better performance in terms of information gain as measured by SPCE and SNMC.
In Figure \ref{fig:spce-CES}  in the paper, we observe that in experiments 0-2, our approach temporarily loses its advantage over RL-BOED due to insufficiently refined particle approximations of the posteriors. However, this edge is regained in subsequent experiments as more information from the posteriors becomes available. Our better design sequences are  also visible in the Wasserstein distance plot presented in Figure \ref{fig:spce-CES}.
Figure \ref{fig:comp_bounds} left shows on the same plot the SPCE and SNMC curves. Without tempering, SMC gains an advantage only in the latter steps 7-9 in terms of information gained, while, in the Wasserstein distance plot presented in Figure \ref{fig:spce-CES},  SMC shows better performance from the start.
A possible explanation is that, as shown in Figure \ref{fig:ces_all_w} below,  RL-BOED performs better on parameter $\rho$ at the expense of sacrificing precision on the others. Overall the Wasserstein distance for all parameters remains in favor of our methods  but it may be that a better precision on $\rho$ leads to a slightly higher information gain (Figure \ref{fig:comp_bounds} left).
}

In our current tempering implementation, the Markov kernel is fixed to a standard Metropolis-Hastings scheme for all steps.
It is out of the scope of this paper but possible directions for improvement include
using more sophisticated kernels, such as Langevin or Hamiltonian Monte Carlo moves, as suggested in the {\it Tuning parameters} section p.1591 of \citet{dai2022invitation} and in references therein. More generally, a number of recommendations, as reviewed in \cite{dai2022invitation}, have been reported as efficient and could be investigated.

\begin{figure}[h!]
  \centering
      \resizebox{\textwidth}{!}{%
  \begin{subfigure}{0.493\textwidth}
    \centering
    \includegraphics[width=\linewidth]{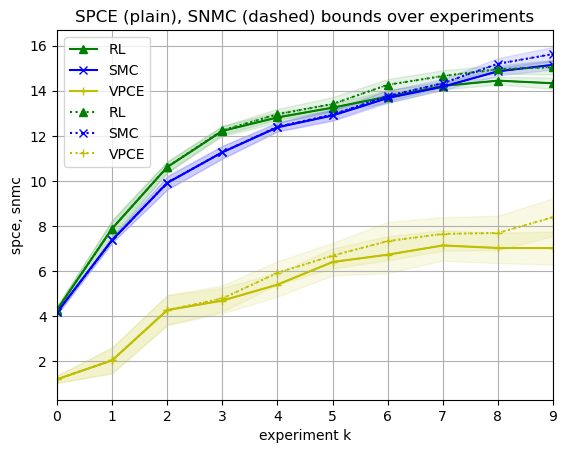}
  \end{subfigure}%
  \hfill
  \begin{subfigure}{0.49\textwidth}
    \centering
    \includegraphics[width=\linewidth]{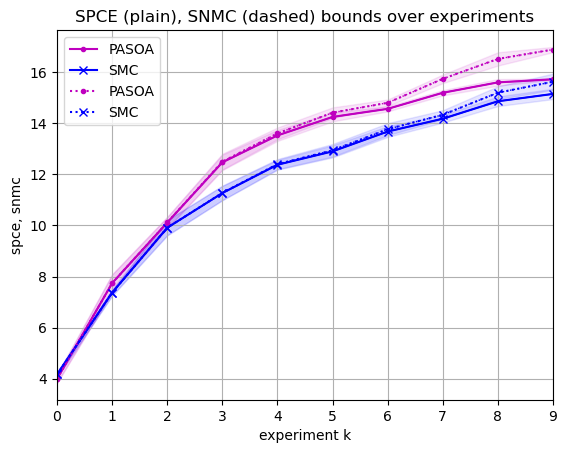}
  \end{subfigure}%
  }
  \caption{CES example. Median and standard error over 100 tollouts, with respect to the number of experiments $k$. The [SPCE, SNMC] intervals containing the totel EIG are plot, respectively with plain (SPCE lower bound) and dashed (SNMC upper bound) lines. Left:  SMC (blue) vs RL-BOED (green) and VPCE (yellow). Right: SMC (blue) vs PASOA (red).}
  \label{fig:comp_bounds}
\end{figure}

\begin{figure}[h!]
\resizebox{\textwidth}{!}{%
\begin{tabular}{ccc}
    \includegraphics[width=0.33\linewidth]{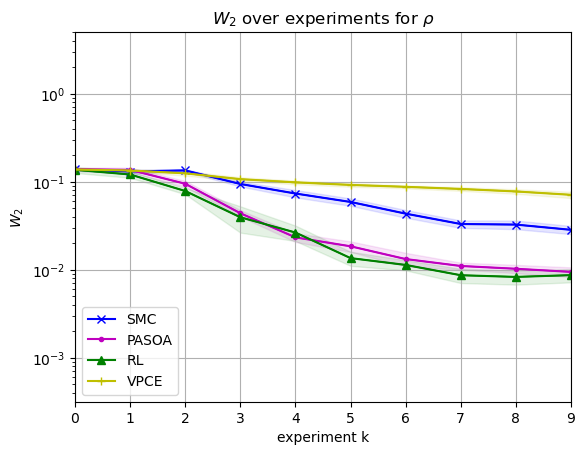} &
  \includegraphics[width=0.33\linewidth]{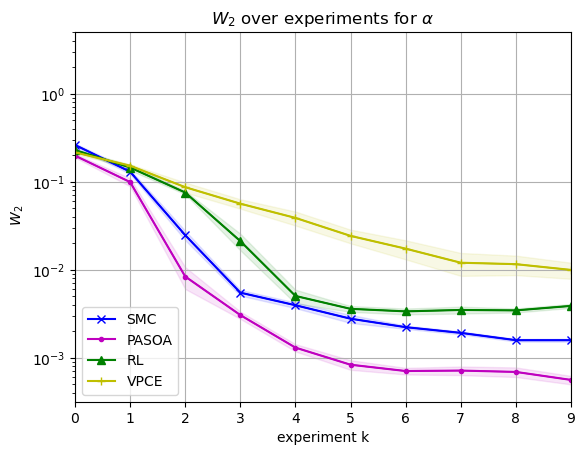} &
    \includegraphics[width=0.33\linewidth]{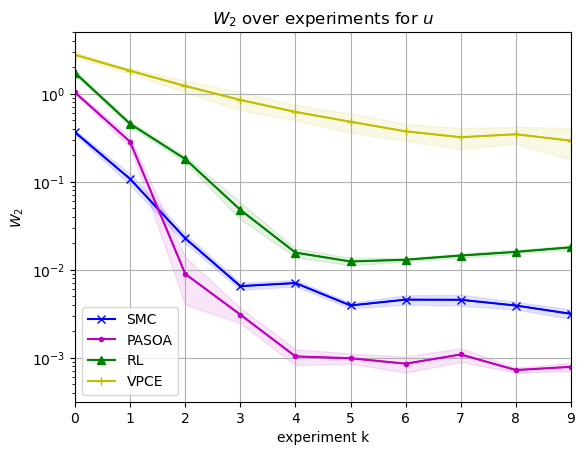}
  \end{tabular}
  }
  \caption{CES example.  Median and standard error of Wasserstein  distances for each  parameter $(\rho, \alphab, u)$  separately.}
  \label{fig:ces_all_w}
\end{figure}

\subsection{Non differentiable examples}

When the model log-likelihood is not differentiable, either because the gradient is not available or difficult to compute, or because the design space is discrete,  the stochastic gradient part of our method cannot be directly applied. However, we can still use the other parts by replacing the optimization step by either an exhaustive argmax, in the case of a finite design space, or by Bayesian optimization \cite{Snoek2012,Hernandez2014,Benassi2011} which does not requires gradients.
This can be seen as an advantage of myopic solutions, which allow such replacements to be easily performed. For each sequential optimization, the search space remains of reasonable size or dimension and does not increases exponentially. This is not the case for other policy-based approach, {\it e.g.} \cite{Foster2021}, which would
involve a challenging high-dimensional Bayesian optimization  of the
policy parameters, or an exhaustive search in an exponentially increasing space with $K$.

To illustrate this situation,  another benchmark example used in \cite{Blau2022,Moffat2020} is the Prey population example. The design is a discrete variable. Instead of using stochastic gradient descent to optimize the $I^{k}_{PCE}$ bound at each step sequentially, we can compute it for every possible design and take the argmax.  The same can be done to adapt VPCE to this discrete design space, while RL-BOED has the advantage to be applicable for both continuous and discrete spaces. Without the gradient part, our approach is similar to that of \citet{Moffat2020} but with an additional tempering, which was already reported to compare favorably to RL-BOED in Figure~4 of \citet{Blau2022}.

%\bibliographystyle{plain}
%\bibliography{biblioBOED}

\end{document}

%% file: preamble.tex
\usepackage[utf8]{inputenc} % allow utf-8 input
\usepackage[T1]{fontenc}    % use 8-bit T1 fonts
\usepackage{hyperref}       % hyperlinks
\hypersetup{
      colorlinks,
      urlcolor={black},
}
\usepackage{url}            % simple URL typesetting
\usepackage{booktabs}       % professional-quality tables
\usepackage{amsfonts}       % blackboard math symbols
\usepackage{nicefrac}       % compact symbols for 1/2, etc.
\usepackage{microtype}      % microtypography

\usepackage{subcaption}
\usepackage[most]{tcolorbox}

\usepackage{tikz,pgfplots}
\usetikzlibrary{decorations.text}
\usetikzlibrary{calc}

\usetikzlibrary{backgrounds,fit,decorations.pathreplacing, angles, quotes, patterns, math}  % TikZ libraries

\usepackage{graphicx}
\usepackage{amssymb,amsmath,enumerate}
\graphicspath{{pics/}{figs/}}
\usepackage{afterpage}
\usepackage{textcomp}

\usepackage{xspace}
\usepackage{bbm}

\usepackage{pifont}
\newcommand{\rcross}{{\color{red}\ding{55}}}
\newcommand{\greencheck}{{\color{green}\checkmark}}

\def\XS{\xspace}
\DeclareMathAlphabet{\mathb}{OML}{cmm}{b}{it}
\def\sbm#1{\ensuremath{\mathb{#1}}}                % Style gras italique (necessite amsmath)
\def\sbmm#1{\ensuremath{\boldsymbol{#1}}}          % Style gras italique (necessite amsmath)
\def\Db{{\sbm{D}}\XS}

\def\Ib{{\sbm{I}}\XS}

\def\alphab      {{\sbmm{\alpha}}\XS}

\def\thetab      {{\sbmm{\theta}}\XS}      \def\Thetab    {{\sbmm{\Theta}}\XS}

\def\xib         {{\sbmm{\xi}}\XS}

\def\PM{\kern0pt^{\textrm{{\scriptsize PM}}}\kern0pt}
\def\MMAP{\kern1pt^{\textrm{{\tiny MMAP}}}\kern-1pt}

\def\Exp{\mathbb{E}} % esperance sur l'espace de la chaine
\newcommand{\wv}{\ensuremath{\mathbf{w}}} % Vecteur des var observees.(obs)
\newcommand{\Xv}{\ensuremath{\mathbf{X}}} % Vecteur des var observees.(obs)
\newcommand{\xv}{\ensuremath{\mathbf{x}}} % Vecteur des var observees.(obs)
\newcommand{\Yv}{\ensuremath{\mathbf{Y}}} % Vecteur des var observees.(obs)
\newcommand{\yv}{\ensuremath{\mathbf{y}}} % Vecteur des var observees.(obs)
\def\Rset{\mathbb{R}} % reels
\def\wp1{\mathrm{w.p.} 1}  % with respect to
\makeatletter

\newtheorem{lem}{\protect\lemmaname}
\newtheorem{proposition}{Proposition}
\newtheorem{theorem}{Theorem}

\makeatother

\usepackage{babel}
\providecommand{\definitionname}{Definition}
\providecommand{\lemmaname}{Lemma}
\providecommand{\remarkname}{Remark}
\newtheorem{assumption}{{\sf A}\hspace{-3pt}}